\documentclass[dvips,sts]{imsart}

\RequirePackage[OT1]{fontenc}
\usepackage{amsthm, amsmath, amssymb,natbib,stmaryrd,bbm}
\usepackage{subfig,url}
\usepackage{graphicx,color}
\RequirePackage[colorlinks,citecolor=blue,urlcolor=blue,breaklinks]{hyperref}

\startlocaldefs
\numberwithin{equation}{section}
\theoremstyle{plain}

\endlocaldefs

\newcommand{\mysec}[1]{Section~\ref{sec:#1}}

\newcommand{\BEAS}{\begin{eqnarray*}}
\newcommand{\EEAS}{\end{eqnarray*}}
\newcommand{\BIT}{\begin{itemize}}
\newcommand{\EIT}{\end{itemize}}
\newcommand{\BNUM}{\begin{enumerate}}
\newcommand{\ENUM}{\end{enumerate}}

\def\Ab{{\mathbf A}}

\def\Hcal{\mathcal{H}}

\def\Tcal{\mathcal{T}}
\def\Integer{{\mathbb{N}}}
\def\Acal{\mathcal{A}}
\def\Pcal{\mathcal{P}}
\def\Gcal{\mathcal{G}}
\def\Xc{\mathcal{X}}
\def\Yc{\mathcal{Y}}
\def\Gc{\mathcal{G}}

\def\Bcal{\mathcal{B}}
\def\Real{{\mathbb{R}}}

\def\R{{\mathbb{R}}}
\newcommand{\SET}[1]{\llbracket 1; #1\rrbracket}
\def\defin{\triangleq}
\def\wb{{\mathbf w}}
\def\fb{{\mathbf f}}
\def\gb{{\mathbf g}}
\def\Wb{{\mathbf W}}
\def\J{{J}}
\def\fro{{\scriptscriptstyle \mathrm{F}}}

\def\vb{{\mathbf v}}
\def\ub{{\mathbf u}}

\def\yb{{\mathbf y}}
\def\vw{{\mathbf w}}
\def\Kb{{\mathbf K}}
\def\vv{{\mathbf v}}
\def\vx{{\mathbf x}}

\def\Xb{{\mathbf X}}
\def\alphab{{\boldsymbol\alpha}}
\def\xb{{\mathbf x}}
\def\Db{{\mathbf D}}
\def\db{{\mathbf d}}
\DeclareMathOperator*{\argmin}{arg\,min}
\def\sign{{\mathrm{sign}}}

\newcommand{\BEQ}{\begin{equation}}
\newcommand{\EEQ}{\end{equation}}
\newcommand{\yh}{\hat{y}}

\begin{document}

\begin{frontmatter}
\title{Structured sparsity through convex optimization}
\runtitle{Structured sparsity through convex optimization}

\begin{aug}
\author{\fnms{Francis} \snm{Bach}\ead[label=e1]{francis.bach@inria.fr}},
\author{\fnms{Rodolphe} \snm{Jenatton}\ead[label=e2]{rodolphe.jenatton@inria.fr}},
\author{\fnms{Julien} \snm{Mairal}\ead[label=e3]{julien@stat.berkeley.edu}}
\and
\author{\fnms{Guillaume} \snm{Obozinski}\ead[label=e4]{guillaume.obozinski@inria.fr}}

 \runauthor{Bach et al.}

\affiliation{INRIA and University of California, Berkeley}

\address{Sierra Project-Team, Laboratoire d'Informatique de l'Ecole Normale Sup\'erieure, Paris, France \printead{e1,e2,e4}.}

\address{Department of Statistics, University of California, Berkeley, USA \printead{e3}.}
\end{aug}

\begin{abstract}
 Sparse estimation methods are aimed at using or obtaining parsimonious representations of data or models. While naturally cast as a combinatorial optimization problem, variable or feature selection admits a convex relaxation through the regularization by the $\ell_1$-norm. In this paper, we consider situations where we are not only interested in sparsity, but where some structural prior knowledge is available as well. We show that the $\ell_1$-norm can then be extended to structured norms built on either disjoint or overlapping groups of variables, leading to a flexible framework that can deal with various structures. 
 We present applications to unsupervised learning, for structured sparse principal component analysis and hierarchical dictionary learning, and to supervised learning in the context of non-linear variable selection. 
  \end{abstract}

\begin{keyword}
\kwd{Sparsity}
\kwd{Convex optimization}
\end{keyword}

\end{frontmatter}

 \section{Introduction}

The concept of parsimony is central in many scientific domains. In the context of statistics, signal processing or machine learning, it takes the form of variable or feature selection problems, and is commonly used in two situations: First,  to make the model or the prediction more interpretable or cheaper to use, i.e., even if
the underlying problem does not admit sparse solutions, one looks for the best sparse approximation. Second, sparsity  can also be used given prior knowledge that the model should be sparse.

Sparse linear models seek to predict an output by linearly combining a small subset of the features describing the data.
To simultaneously address variable selection and model estimation, $\ell_1$-norm regularization has become a popular tool, 
which benefits both from efficient algorithms~\citep[see, e.g.,][and multiple references therein]{Efron2004,Beck2009, Yuan2010,Bach2011} 
and a well-developed theory for generalization properties and variable selection consistency~\citep{Zhao2006,Wainwright2009,Bickel2009,Zhang2009}.

When regularizing with the $\ell_1$-norm, each variable is selected individually, 
regardless of its position in the input feature vector, 
so that existing relationships and structures
between the variables (e.g., spatial, hierarchical or related to the physics of the problem at hand) are merely disregarded.
However, in many practical situations the estimation can benefit from some type of prior knowledge, 
potentially both for interpretability and to improve predictive performance.

This a priori can take various forms: in neuroimaging based on functional magnetic resonance (fMRI) or
magnetoencephalography (MEG), sets of voxels allowing to discriminate between different brain states are expected to form small localized and connected areas \citep[and references therein]{Gramfort2009, Xiang2009}.
Similarly, in face recognition, as shown in \mysec{sspca}, robustness to occlusions can be increased by considering as features,
sets of pixels that form small convex regions of the faces. 
Again, a plain $\ell_1$-norm regularization fails to encode such specific spatial constraints~\citep{Jenatton2010}.
The same rationale supports the use of \textit{structured sparsity} 
for background subtraction~\citep{Cevher2008,Huang2011,Mairal2011}. 

Another example of the need for higher-order prior knowledge comes from bioinformatics.
Indeed, for the diagnosis of tumors, the profiles of array-based comparative genomic hybridization (arrayCGH) can be used as inputs 
to feed a classifier \citep{Rapaport2008}.
These profiles are characterized by many variables, but only a few observations of such profiles are available, prompting the need for variable selection.
Because of the specific spatial organization of bacterial artificial chromosomes along the genome, 
the set of discriminative features is expected to consist of specific contiguous patterns. 
Using this prior knowledge in addition to standard sparsity leads to improvement in classification accuracy \citep{Rapaport2008}.  
In the context of multi-task regression, a problem of interest in genetics is to find a mapping between 
a small subset of loci presenting single nucleotide polymorphisms (SNP's) that have a phenotypic impact on a given family of genes \citep{Kim2009}.
This target family of genes has its own structure, where some genes share common genetic characteristics, 
so that these genes can be embedded into some underlying hierarchy. 
Exploiting directly this hierarchical information in the regularization term outperforms the unstructured approach with a standard $\ell_1$-norm~\citep{Kim2009}.

These real world examples motivate the need for the design of sparsity-inducing regularization schemes, 
capable of encoding more sophisticated prior knowledge about the expected sparsity patterns. 
As mentioned above, the $\ell_1$-norm corresponds only to a constraint on \textit{cardinality} 
and is oblivious of any other information available about the patterns of nonzero coefficients (``nonzero patterns'' or ``supports'') induced in the solution, 
since they are all theoretically possible. 
In this paper, we consider a family of sparsity-inducing norms that can address a large variety of structured sparse problems: a simple change of norm will induce new ways of selecting variables; moreover, as shown in \mysec{prox} and \mysec{theory}, algorithms to obtain estimators (e.g., convex optimization methods) and theoretical analyses are easily extended in many situations. As shown in \mysec{structsparsity}, the norms we introduce generalize traditional ``group $\ell_1$-norms'', that have been popular for selecting variables organized in non-overlapping groups~\citep{Turlach2005,Yuan2006,Roth2008,Huang2010}. Other families for different types of structures are presented in Section~\ref{sec:related}.

The paper is organized as follows: we first review in \mysec{l1} classical $\ell_1$-norm regularization in supervised contexts. We then introduce several families of norms in \mysec{structsparsity}, and present applications to unsupervised learning in \mysec{unsup}, namely for sparse principal component analysis in \mysec{sspca} and hierarchical dictionary learning in \mysec{hdl}. We briefly show in \mysec{hkl} how these norms can also be used for high-dimensional non-linear variable selection.

\paragraph{Notations.}
Throughout the paper, we shall denote vectors with bold lower case   letters, 
and matrices with bold upper case ones.
For any integer $j$ in the set $\SET{p}\defin\{1,\dots,p\}$, we denote the $j$-th coefficient of a 
$p$-dimen\-sio\-nal vector $\wb \in \Real^p$ by $\wb_j$.
Similarly, for any matrix $\Wb \in \Real^{n\times p}$, we refer to the entry on the $i$-th row and $j$-th column as
$\Wb_{ij}$, for any $(i,j) \in \SET{n}\times\SET{p}$.
We will need to refer to sub-vectors of $\wb \in \Real^p$, and so, for any $\J \subseteq \SET{p}$, 
we denote by $\wb_\J \in \Real^{|\J|}$ the vector consisting of the entries of $\wb$ indexed by $\J$.
Likewise, for any $I \subseteq \SET{n}$, $\J \subseteq \SET{p}$, we denote by 
$\Wb_{I\J} \in \Real^{|I| \times |\J|}$ the sub-matrix of $\Wb$ formed by the rows (respectively the columns)
indexed by $I$ (respectively by $\J$).
We extensively manipulate norms in this paper. We thus define the $\ell_q$-norm for any vector $\wb \in \Real^p$
by
$
\|\wb\|_q^q\defin  \sum_{j=1}^p |\wb_j|^q  \ \text{for}\ q\in[1,\infty),\quad \text{and}\quad
\|\wb\|_\infty\defin \max_{j\in\SET{p}}|\wb_j|.
$
For $q \in (0,1)$, we extend the definition above to $\ell_q$ pseudo-norms.
Finally,  for any matrix $\Wb \in \Real^{n\times p}$, we define the Frobenius norm of $\Wb$ by
$
\|\Wb\|_\fro^2 \defin  \sum_{i=1}^n\sum_{j=1}^p \! \Wb_{ij}^2  .
$

\section{Unstructured sparsity via the $\ell_1$-norm}
\label{sec:l1}

Regularizing by the $\ell_1$-norm has been a topic of intensive research over the last decade. 
This line of work has witnessed the development of nice 
theoretical frameworks~\citep{Tibshirani1996,  Chen1998, Mallat1999, Tropp2004, Tropp2006, Zhao2006, Zou2006, Wainwright2009, Bickel2009, Zhang2009, Negahban2009} and the emergence of many efficient
algorithms~\citep{Efron2004, Nesterov2007, Friedman2007, Wu2008, Beck2009, Wright2009, Needell2009, Yuan2010a}. 
Moreover, this methodology has found quite a few applications, notably in compressed sensing~\citep{Candes2005}, for the estimation of the structure of graphical models~\citep{Meinshausen2006} or
for several reconstruction tasks involving natural images~\citep[e.g., see][for a review]{Mairal2010b}.
In this section, we focus on supervised learning and 
present the traditional estimation problems associated with sparsity-inducing norms such as the $\ell_1$-norm~(see \mysec{unsup} for unsupervised learning).

In supervised learning, we predict (typically one-dimensional) outputs $y$ in~$\Yc$ from observations $\vx$ in $\Xc$;
these observations are usually represented by $p$-dimen\-sio\-nal vectors with $\Xc=\R^p$. 
M-estimation and in particular regularized empirical risk minimization are well suited to this setting.
Indeed, given $n$ pairs of data points
$
\{ (\vx^{(i)},y^{(i)}) \in \R^p\! \times\! \Yc;\ i\!=\!1,\dots,n\}
$,
we consider the estimators solving the following form of convex optimization problem
\begin{equation}
\label{eq:formulation}
   \min_{\vw \in \R^p} \frac{1}{n}\sum_{i=1}^n\ell(y^{(i)}, {\vw}^\top\vx^{(i)}) + \lambda \Omega(\vw),  \end{equation}
where $\ell$ is a loss function and $\Omega: \R^p \to \R$ is a sparsity-inducing---typically nonsmooth and non-Euclidean---norm.
Typical examples of differentiable loss functions are the square loss for least squares regression,
i.e., $\ell(y,\yh) = \frac{1}{2}(y-\yh)^2$ with~$y$ in $\R$,
and the logistic loss $\ell(y,\yh) = \log(1+e^{-y\yh})$ for logistic regression, with $y$ in $\{-1,1\}$.
We refer the readers to \cite{Shawe-Taylor2004} and to \citet{hastie} for more complete descriptions of loss functions.

Within the context of least-squares regression, $\ell_1$-norm regularization is known as the Lasso~\citep{Tibshirani1996} in statistics and as basis pursuit in signal processing~\citep{Chen1998}.
For the Lasso, formulation (\ref{eq:formulation}) takes the form
\begin{equation}\label{intro:eq:lasso}
\min_{\wb \in \Real^p} \frac{1}{2n} \|\yb - \Xb\wb\|_2^2 + \lambda \|\wb\|_1,
\end{equation}
and, equivalently, basis pursuit can be written\footnote{Note that the formulations which are typically encountered in signal processing are either $\min_\alphab \|\alphab\|_1\: \text{s.t.}\: \xb=\Db\alphab$, which corresponds to the limiting case of Eq.~(\ref{intro:eq:basispursuit}) where $\lambda \rightarrow 0$ and $\xb$ is in the span of the dictionary $\Db$, or $\min_\alphab \|\alphab\|_1\: \text{s.t.}\: \|\xb-\Db\alphab\|_2 \leq \eta$ which is a constrained counterpart of Eq.~(\ref{intro:eq:basispursuit}) leading to the same set of solutions (see the explanation following  Eq.~(\ref{intro:eq:minf_c})).
}
\begin{equation}\label{intro:eq:basispursuit}
\min_{\alphab \in \Real^p} \frac{1}{2} \|\xb - \Db\alphab\|_2^2 + \lambda \|\alphab\|_1.
\end{equation}
These two equations are obviously identical but we write them both 
to show the correspondence between notations used in statistics and in signal processing.
In statistical notations, we will use $\Xb\in\Real^{n\times p}$ to denote a set of $n$ observations described by $p$ variables (covariates), 
while $\yb \in \Real^n$ represents the corresponding set of $n$ targets (responses) that we try to predict. For instance, $\yb$ may have discrete entries in the context of classification.
With notations of signal processing, we will consider an $m$-dimensional signal $\xb \in \Real^m$ that we express as a linear combination of $p$ dictionary elements composing the 
dictionary $\Db \defin [\db^1,\dots,\db^p] \in \Real^{m\times p}$.
While the design matrix $\Xb$ is usually assumed fixed and given beforehand, we shall see in Section~\ref{sec:unsup} 
that the dictionary $\Db$ may correspond either to some pre-defined basis \citep[e.g., see][for wavelet bases]{Mallat1999} 
or to a representation that is actually \emph{learned} as well~\citep{Olshausen1996}.

\paragraph{Geometric intuitions for the $\ell_1$-norm ball.}
While we consider in~(\ref{eq:formulation}) a regularized formulation, we could have considered an equivalent \emph{constrained} problem of the form
\begin{equation}\label{intro:eq:minf_c}
\min_{\wb \in \Real^p} \frac{1}{n}\sum_{i=1}^n\ell(y^{(i)}, {\vw}^\top\vx^{(i)})  \quad \text{such that}\quad  \Omega(\wb) \leq \mu, 
\end{equation}
for some $\mu \in \Real_+$: 
It is indeed the case that the solutions to problem (\ref{intro:eq:minf_c}) obtained when varying $\mu$ is the same as the solutions to problem~(\ref{eq:formulation}), 
for some of $\lambda_\mu$ depending on $\mu$~\citep[e.g., see Section 3.2 in][]{Borwein2006}.

\begin{figure}[ht]
 \centering
   \subfloat[$\ell_2$-norm ball]{\includegraphics[width=0.3\linewidth]{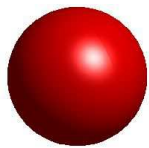}} \hspace*{2cm}   \subfloat[$\ell_1$-norm ball]{\includegraphics[width=0.3\linewidth]{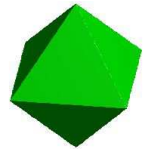}}
\caption{Comparison between the $\ell_2$-norm and $\ell_1$-norm balls in three dimensions, respectively on the left and right figures.
The $\ell_1$-norm ball presents some singular points located along the axes of $\Real^3$ and along the three axis-aligned planes going through the origin.}\label{intro:fig:l1l2balls}
\end{figure}

At optimality, the opposite of the gradient of $f: \wb \mapsto  \frac{1}{n}\sum_{i=1}^n\ell(y^{(i)}, {\vw}^\top\vx^{(i)})$ evaluated at any solution~$\hat{\wb}$ of~(\ref{intro:eq:minf_c}) must belong to the normal cone to $\Bcal=\{\wb \in \Real^p;\ \Omega(\wb) \leq \mu\}$
at~$\hat{\wb}$ \citep{Borwein2006}. In other words, 
for sufficiently small values of $\mu$ (i.e., ensuring that the constraint is active) 
the level set of~$f$ for the value $f(\hat{\wb})$ is tangent to $\Bcal$.
As a consequence, important properties of the solutions $\hat{\wb}$ follow from the geometry of the ball $\Bcal$.
If $\Omega$ is taken to be the $\ell_2$-norm, then the resulting ball $\Bcal$ is the standard, isotropic, ``round'' ball that does not favor any specific direction of the space.  
On the other hand, when $\Omega$ is the $\ell_1$-norm, $\Bcal$ corresponds to a diamond-shaped pattern in two dimensions, and to a double pyramid in three dimensions.
In particular, $\Bcal$ is anisotropic and exhibits some singular points due to the non-smoothness of $\Omega$. 
Since these singular points are located along axis-aligned linear subspaces in~$\Real^p$, if the level set of $f$ with the smallest feasible value is tangent to $\Bcal$ at one of those points, sparse solutions are obtained.
We display on Figure~\ref{intro:fig:l1l2balls} the balls $\Bcal$ for both the $\ell_1$- and $\ell_2$-norms. See \mysec{structsparsity} and Figure~\ref{intro:fig:groupl1l2balls} for extensions to structured norms.

\section{Structured Sparsity-Inducing Norms}
\label{sec:structsparsity}
In this section, we consider structured sparsity-inducing norms that induce estimated vectors that are not only sparse, as for the $\ell_1$-norm, but whose support also displays some
structure known a priori that reflects potential relationships between the variables. 
\subsection{Sparsity-Inducing Norms with Disjoint Groups of Variables}\label{subsec:grouplasso}

\begin{figure}[ht]
 \centering
   \subfloat[$\ell_1/\ell_2$-norm ball without overlaps: $\Omega(\wb)=\|\wb_{\{1,2\}}\|_2+|\wb_3|$]{\hspace*{.5cm}\includegraphics[width=0.3\linewidth]{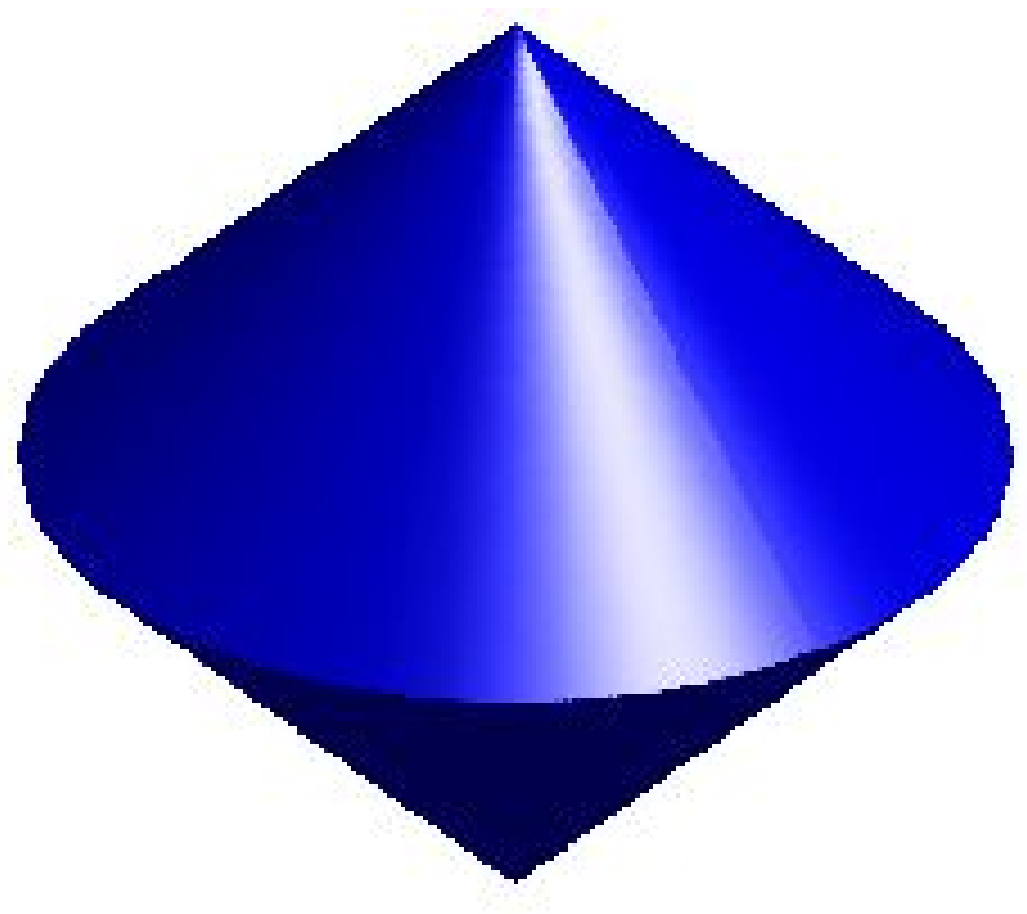}\label{intro:subfig:groupl1l2ball}\hspace*{.5cm}} \hfill
   \subfloat[$\ell_1/\ell_2$-norm ball with overlaps: $\Omega(\wb)=\|\wb_{\{1,2,3\}}\|_2+|\wb_1|+|\wb_2|$]{\hspace*{.5cm}\includegraphics[width=0.3\linewidth]{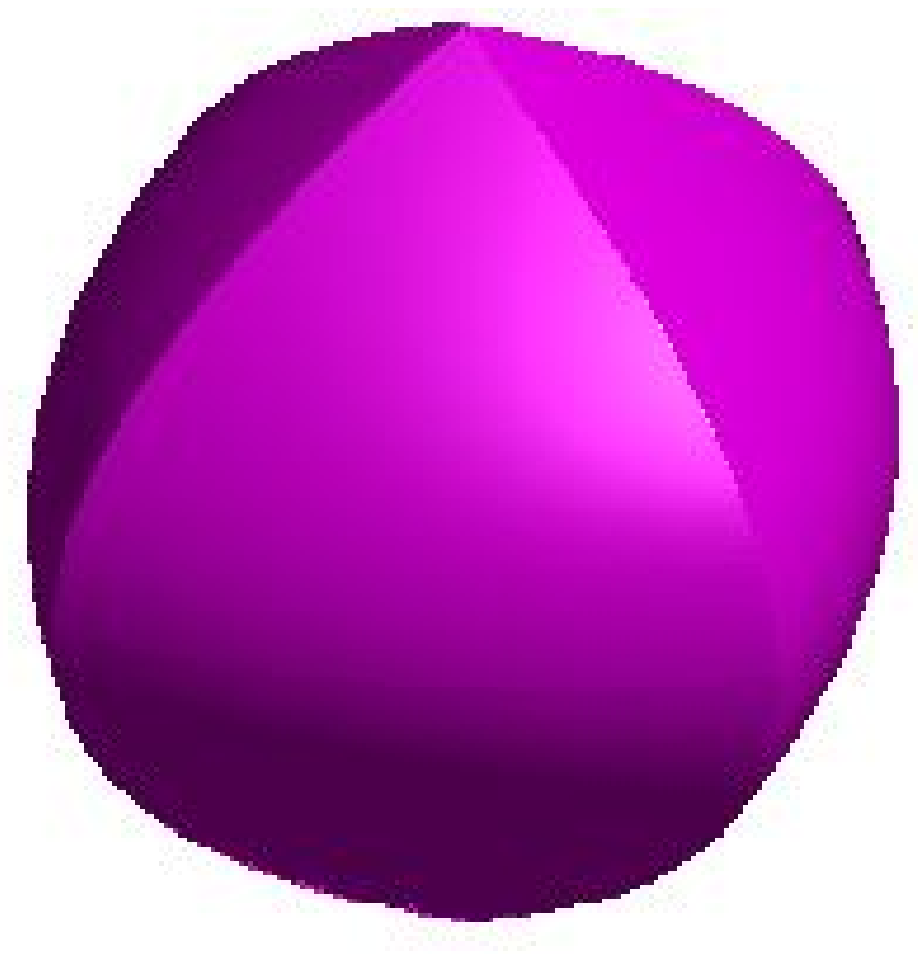}\label{intro:subfig:overl1l2ball}\hspace*{.5cm}}
\caption{Comparison between two mixed $\ell_1/\ell_2$-norm balls in three dimensions (the first two directions are horizontal, the third one is vertical), without and with overlapping groups of variables, 
respectively on the left and right figures. The singular points appearing on these balls describe the sparsity-inducing behavior of the underlying norms $\Omega$.}\label{intro:fig:groupl1l2balls}
\end{figure}
The most natural form of structured sparsity is arguably \emph{group sparsity}, matching the a priori knowledge that pre-specified disjoint blocks of variables should be selected or ignored simultaneously.
In that case, if $\Gcal$ is a collection of groups of variables, forming a partition of $\SET{p}$, and $d_g$ is a positive scalar weight indexed by group $g$, we define $\Omega$ as
\begin{equation}\label{intro:eq:omega}
\Omega(\wb) = \sum_{g\in\Gcal} d_g \|\wb_g\|_q\quad \text{for any}\ q \in (1,\infty].
\end{equation}
This norm is usually referred to as a mixed $\ell_1/\ell_q$-norm, and in practice, popular choices for $q$ are $\{2,\infty\}$. As desired, regularizing with $\Omega$ leads variables in the same group to be selected or set to zero simultaneously (see Figure~\ref{intro:fig:groupl1l2balls} for a geometric interpretation).
In the context of least-squares regression, this regularization is known as the group Lasso~\citep{Turlach2005, Yuan2006}.
It has been shown to improve the
prediction performance and/or interpretability of the learned models when the block structure is relevant~\citep{Roth2008,Stojnic2009,Lounici2009,Huang2010}.
Moreover, applications of this regularization scheme arise also in the context of multi-task learning~\citep{Obozinski2009,Quattoni2009,Liu2009a} to account for features shared across tasks,
and multiple kernel learning~\citep{Bach2008a} for the selection of different kernels (see also \mysec{hkl}).

\paragraph{Choice of the weights.}
When the groups vary significantly in size, results can be improved, in particular under high-dimensional scaling, by an
appropriate choice of the weights $d_g$ which compensate for the discrepancies of sizes between groups. It is difficult to provide a unique choice for the weights. In general, they depend on $q$ and on the type of consistency desired. We refer the reader to \citet{Yuan2006,Bach2008a,Obozinski2011Group,Lounici2011} for general discussions.

It might seem that the case of groups that overlap would be unnecessarily complex. It turns out, in reality, that appropriate collections of overlapping groups allow to encode quite interesting forms of structured sparsity. In fact, the idea of constructing sparsity-inducing norms from overlapping groups will be key. We present two different constructions based on overlapping groups of variables that are essentially complementary of each other in Sections~\ref{sec:inter-closed} and~\ref{sec:union-closed}.

\subsection{Sparsity-Inducing Norms with Overlapping Groups of Variables}
\label{sec:inter-closed}
In this section, we consider a direct extension of the norm introduced in the previous section to the case of overlapping groups; we give an informal overview of the structures that it can encode and examples of relevant applied settings.
For more details see \citet{Jenatton2009}.

Starting from the definition of $\Omega$ in Eq.~(\ref{intro:eq:omega}),
it is natural to study what happens when the set of groups $\Gcal$ is allowed to contain elements that overlap.
In fact, and as shown by~\citet{Jenatton2009}, the sparsity-inducing behavior of~$\Omega$ remains the same:
when regularizing by $\Omega$, some entire groups of variables~$g$ in $\Gcal$ are set to zero.
This is reflected in the set of non-smooth extreme points of the unit ball of the norm represented on Figure~\ref{intro:fig:groupl1l2balls}-\subref{intro:subfig:overl1l2ball}.
While the resulting patterns of nonzero variables---also referred to as \emph{supports}, or \emph{nonzero patterns}---were obvious in the non-overlapping case, 
it is interesting to understand here the relationship that ties together the set of groups $\Gcal$ and its associated set of possible nonzero patterns.
Let us denote by $\Pcal$ the latter set. For any norm of the form~(\ref{intro:eq:omega}), it is still the case that variables belonging to a given group are encouraged to be set simultaneously to zero; as a result, the possible zero patterns
for solutions of~(\ref{eq:formulation}) are  obtained by forming unions of the basic groups, which means that the possible supports are obtained by taking the intersection of a certain number of complements of the basic groups.

Moreover, under mild conditions \citep{Jenatton2009},
given any \emph{intersection-closed}\footnote{A set $\Acal$ is said to be intersection-closed, 
if for any $k \in \Integer$, and for any $(a_1,\dots,a_k) \in \Acal^k$, we have $\bigcap_{i=1}^k \! a_i \in \Acal$.} 
family of patterns $\Pcal$ of variables (see examples below),
it is possible to build an ad-hoc set of groups $\Gcal$---and hence, a regularization norm~$\Omega$---that enforces 
the support of the solutions of~(\ref{eq:formulation}) to belong to $\Pcal$.

These properties 
make it possible to \emph{design} norms that are adapted to the structure of the problem at hand, which we now illustrate with a few examples.

\begin{figure}[!h]
\begin{center}
\includegraphics[scale=.4]{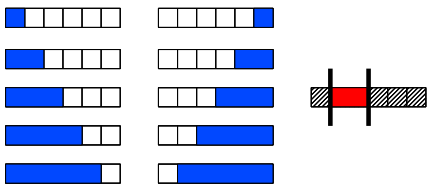}
\end{center}
\caption{ (Left) The set of blue groups to penalize in order to select contiguous patterns in a sequence.
(Right) In red, an example of such a nonzero pattern with its corresponding zero pattern (hatched area). } 
\label{intro:fig:sequence}
 \end{figure}

\paragraph{One-dimensional interval pattern.}
Given $p$ variables organized in a sequence, using the set of groups $\Gcal$ of Figure~\ref{intro:fig:sequence}, it is only possible to select \emph{contiguous nonzero patterns}. In this case, we have $|\Gcal|=O(p)$.
Imposing the contiguity of the nonzero patterns can be relevant in the context of variable forming time series, or for the diagnosis of tumors, based on the profiles of CGH arrays \citep{Rapaport2008}, since a bacterial artificial chromosome will be inserted as a single continuous block into the genome. 

\begin{figure}[!h]
\begin{center}
\includegraphics[width=\linewidth]{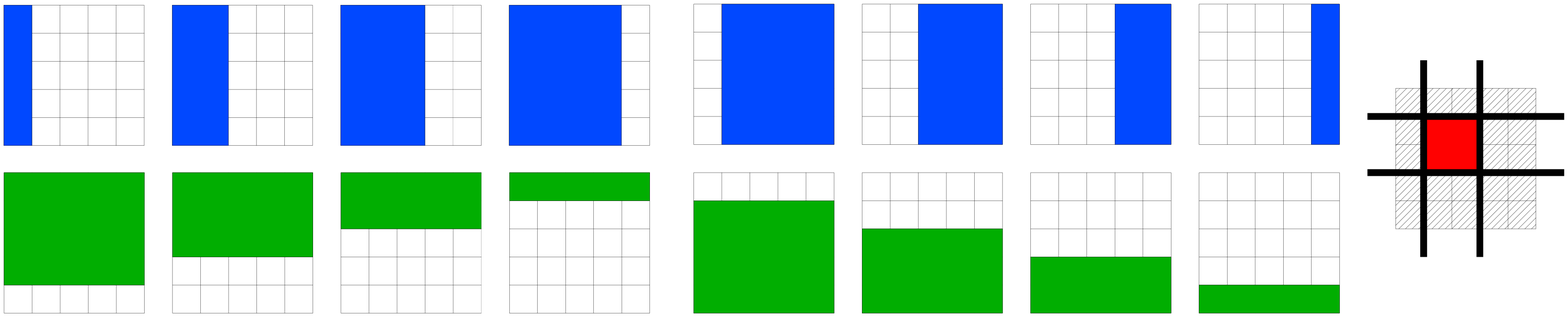} \\[.5cm]
\includegraphics[scale=.4]{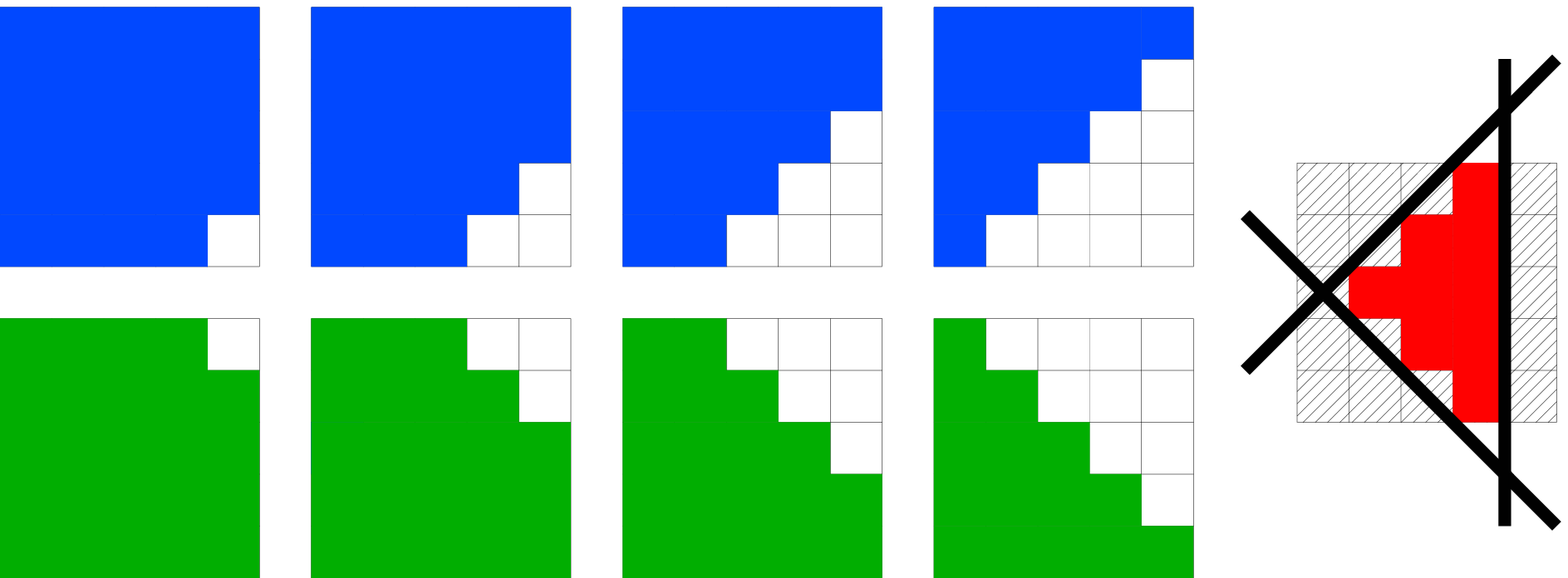}
\end{center}

\caption{ Top: Vertical and horizontal groups: (Left) the set of blue and green groups to penalize in order to select
rectangles. (Right) In red, an example of nonzero pattern recovered in this setting, with its corresponding zero pattern (hatched area). Bottom: Groups with $\pm \pi/4$ orientations: (Left) the set of blue and green groups with their (not displayed) complements to penalize in order to select
diamond-shaped patterns. } 
\label{intro:fig:axis-aligned}
\end{figure}

\paragraph{Two-dimensional convex support.}
Similarly, assume now that the $p$ variables are organized on a two-dimensional grid.
To constrain the allowed supports $\Pcal$ to be the set of all rectangles on this grid, 
a possible set of groups $\Gcal$ to consider is represented in the top of Figure~\ref{intro:fig:axis-aligned}.
This set is relatively small since $|\Gcal|=O(\sqrt{p})$. Groups corresponding to half-planes with additional orientations (see Figure~\ref{intro:fig:axis-aligned} bottom) may be added to ``carve out" more general convex patterns. See an illustration in \mysec{sspca}.

\paragraph{Two-dimensional block structures on a grid.} Using sparsity-inducing regularizations built upon groups which are composed of variables together with their spatial neighbors leads to 
 good performances for background subtraction~\citep{Cevher2008,Baraniuk2008,Huang2011,Mairal2011},
 topographic dictionary learning~\citep{Kavukcuoglu2009, Mairal2011}, and wavelet-based denoising~\citep{Rao2011}.  
 
\begin{figure}[hbtp!]
   \centering
   \includegraphics[width=0.6\textwidth]{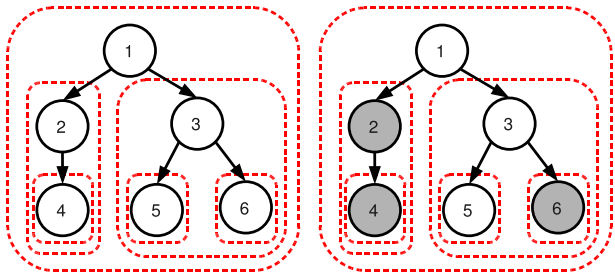}
   \caption{Left:  set of groups $\Gcal$ (dashed contours in red) corresponding to the tree $\mathcal{T}$ with $p=6$ nodes represented by black circles.
   Right: example of a sparsity pattern induced by the tree-structured norm corresponding to $\Gcal$: the groups $\{2,4\},\{4\}$ and $\{6\}$ are set to zero, so that the corresponding nodes (in gray) that form subtrees of $\mathcal{T}$ are removed.
   The remaining nonzero variables $\{1,3,5\}$ form a rooted and connected subtree of $\mathcal{T}$.
   This sparsity pattern obeys the two following equivalent rules: (i) if a node is selected, the same goes for all its ancestors. (ii) if a node is not selected, then its descendant are not selected.
}\label{intro:fig:treegroups}
\end{figure}

\paragraph{Hierarchical structure.}
A fourth interesting example assumes that the variables are organized in a hierarchy. Precisely, 
we assume that the $p$ variables can be assigned to the nodes of a tree $\Tcal$ (or a forest of trees), and that a given variable may be selected 
 only if all its ancestors in $\Tcal$ have already been selected.
This hierarchical rule is exactly respected when using the family of groups displayed on Figure~\ref{intro:fig:treegroups}.
The corresponding penalty was first used by~\citet{Zhao2009}; one of it simplest instance in the context of regression is the sparse group Lasso ~\citep{sprechmann2010collaborative,friedman2010note}; 
it has found numerous applications, for instance, 
 wavelet-based denoising~\citep{Zhao2009, Baraniuk2008,Huang2011,Jenatton2010b},
 hierarchical dictionary learning for both topic modelling and image restoration~\citep{Jenatton2010b},
 log-linear models for the selection of potential orders~\citep{Schmidt2010}, 
 bioinformatics, to exploit the tree structure of gene networks for multi-task regression~\citep{Kim2009},
 and multi-scale mining of fMRI data for the prediction of simple cognitive tasks~\citep{Jenatton2011a}.

\paragraph{Extensions.}
Possible choices for the sets of groups $\Gcal$ are not limited to the aforementioned examples: more complicated topologies can be considered, for example three-dimensional spaces discretized in cubes or spherical volumes discretized in slices (see an application to neuroimaging by \citet{Varoquaux2010a}), and more complicated hierarchical structures based on directed acyclic graphs can be encoded as further developed in \mysec{hkl}.

\paragraph{Choice of the weights.}
The choice of the weights $d_g$ is significantly more important in the overlapping case both theoretically and in practice. In addition to compensating for the discrepancy in group sizes, the weights additionally have to make up for the potential over-penalization of parameters contained in a larger number of groups. For the case of one-dimensional
interval patterns, \cite{Jenatton2009} showed that it was more efficient  in practice to actually weight each individual coefficient \emph{inside} of a group as opposed to weighting the group globally.

\subsection{Norms for Overlapping Groups: a Latent Variable Formulation}
\label{sec:union-closed}
The family of norms defined in Eq.~(\ref{intro:eq:omega}) is adapted to \emph{intersection-closed}
sets of nonzero patterns. However, some applications exhibit structures that can be more naturally modelled by \emph{union-closed} families of supports.
This idea was introduced by \citet{Jacob2009} and \citet{Obozinski2011Group} who, given a set of groups $\Gc$, proposed the following norm
\begin{equation}\label{eq:omegajacob}
\Omega_{\text{union}}(\vw) \defin \min_{ \vv \in \R^{p\times |\Gc|}  } \sum_{ g \in \Gc } d_g \| \vv^g \|_q\quad\text{such that}\ 
\begin{cases}
&\sum_{ g\in\Gc }\vv^g = \vw,\\
&\forall g\in\Gc,\ \vv^g_j = 0\ \text{if}\ j \notin g,
\end{cases}
\end{equation}
where again $d_g$ is a positive scalar weight associated with group $g$.

The norm we just defined provides a different generalization of the $\ell_1/\ell_q$-norm to the case of overlapping groups than the norm presented in \mysec{inter-closed}.
In fact, it is easy to see that solving Eq.~(\ref{eq:formulation}) with the norm $\Omega_{\text{union}}$ is equivalent to solving
\begin{equation}
\label{eq:expanded}
\min_{(\vv^g \in \R^{|g|})_g \in \Gcal} \sum_{i=1}^n \ell \big ( y^{(i)}, \sum_{g \in \Gcal} {\vv^g_g}^\top \vx_g^{(i)}\big ) + \lambda \, \sum_{g \in \Gcal} d_g \|\vv^g\|_q ,
\end{equation}
and setting $\vw=\sum_{g \in \Gcal}\vv^g$.
This last equation shows that using the norm $\Omega_{\text{union}}$ can be interpreted as implicitly duplicating the variables belonging to several groups and regularizing with a weighted $\ell_1/\ell_q$-norm for disjoint groups in the expanded space.
Again in this case a careful choice of the weights is important \citep{Obozinski2011Group}.

This latent variable formulation pushes some of the vectors $\vv^g$ to zero while keeping others with no zero components, hence leading to a vector $\vw$ with a support which is in general the union of the selected groups. 
Interestingly, it can be seen as a convex relaxation of a non-convex penalty encouraging similar sparsity patterns which was introduced by~\cite{Huang2011} and which we present in Section~\ref{sec:submod_blockcoding}.

\textit{Graph Lasso.} One type of a priori knowledge commonly encountered takes the form of a graph defined on the set of input variables, which is such that connected variables are more likely to be simultaneously relevant or irrelevant; this type of prior is common in genomics where regulation, co-expression or interaction networks between genes (or their expression level) used as predictors are often available. To favor the selection of neighbors of a selected variable, it is possible to consider the edges of the graph
as groups in the previous formulation~\citep[see][]{Jacob2009,Rao2011}.

\textit{Patterns consisting of a small number of intervals.} A quite similar situation occurs, when one knows a priori---typically for variables forming sequences (times series, strings, polymers)---that the support should consist of a small number of connected subsequences. In that case, one can consider the sets of variables forming connected subsequences (or connected subsequences of length at most $k$) as the overlapping groups~\citep{Obozinski2011Group}.

\begin{figure}[ht]
 \centering
   \subfloat[Unit ball for $\Gcal=\big \{\{1,3\},\{2,3\} \big \}$]
   {\hspace*{.5cm}\includegraphics[width=0.3\linewidth]{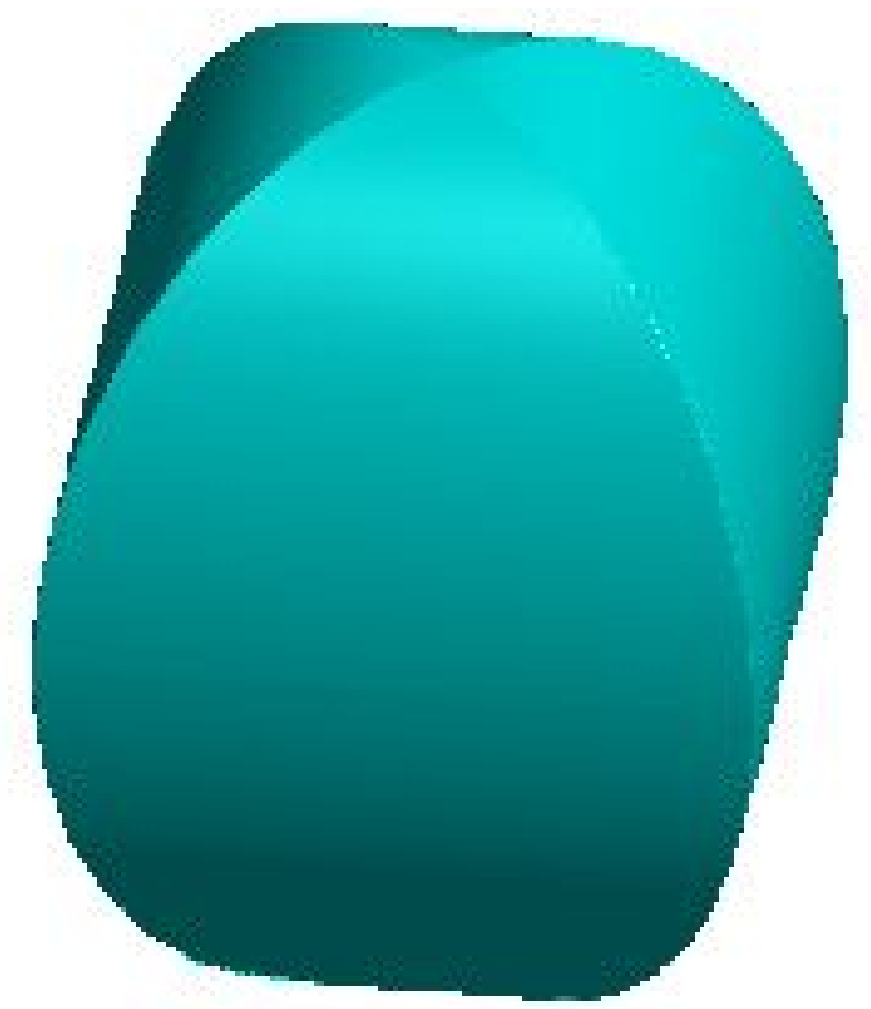}
   \label{subfig:lglball2}\hspace*{.5cm}} \hfill
   \subfloat[$\Gcal=\big \{\{1,3\},\{2,3\},\{1,2\} \big \}$ ]{\hspace*{.5cm}\includegraphics[width=0.3\linewidth]{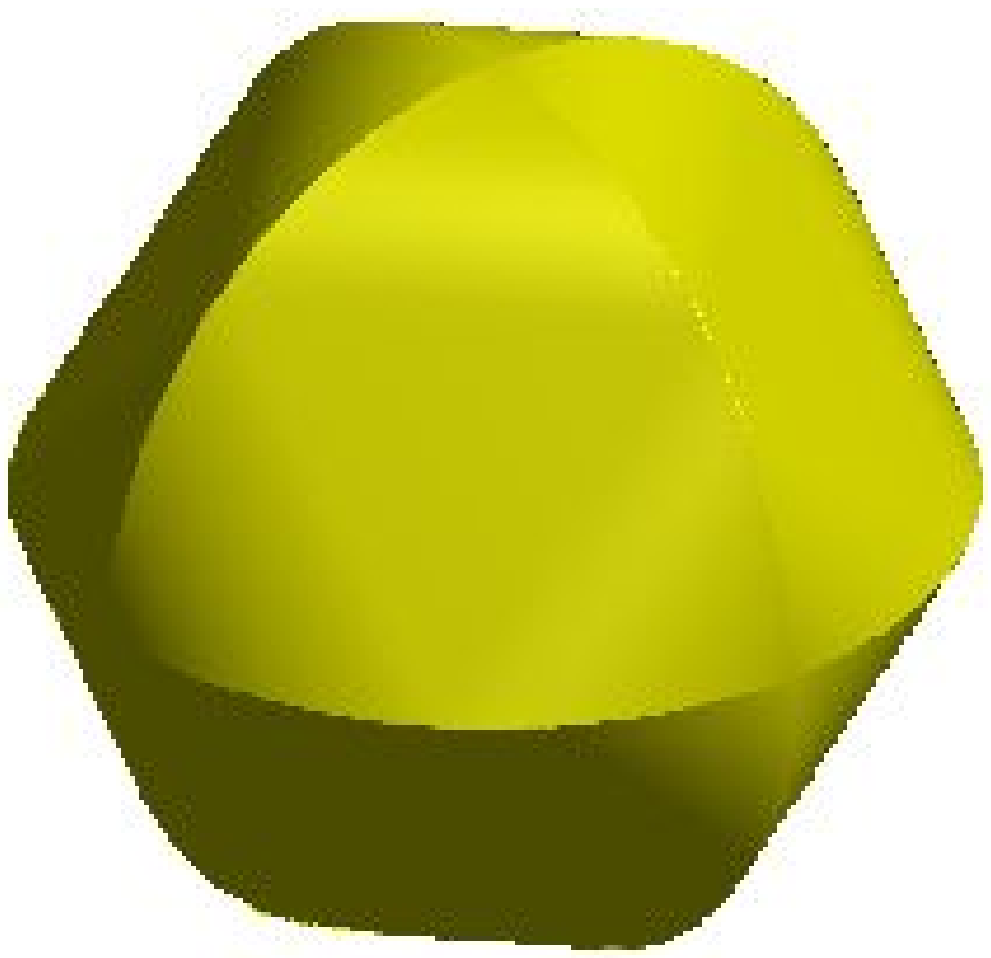}\label{subfig:lglball3}\hspace*{.5cm}}
\caption{Two instances of unit balls of the latent group Lasso regularization  $\Omega_{\text{union}}$  for two or three groups of two variables. Their singular points lie on axis aligned circles, corresponding to each group, and whose convex hull is exactly the unit ball. It should be noted that the ball on the left is quite similar to the one of Fig.~\ref{intro:subfig:overl1l2ball} except that its ``poles" are flatter, hence discouraging the selection of $\vx_3$ without either $\vx_1$ or $\vx_2$.}
\label{fig:lgl_balls}
\end{figure}

\subsection{Related Approaches to Structured Sparsity}
  \label{sec:related}

 \paragraph{Norm design through submodular functions.}
 \label{sec:submod_blockcoding}
 
 Another approach to structured sparsity relies on submodular analysis~\citep{Bach2010a}.
Starting from a non-de\-crea\-sing, submodular\footnote{Let $S$ be a finite set. A function $F: 2^S \rightarrow \Real$ is said to be submodular if for any subset $A,B \subseteq S$, 
we have the inequality $F(A\cap B)+F(A\cup B) \leq F(A) + F(B)$; see~\citet{submodular_tutorial} and references therein.}
 set-function $F$ of the supports of the parameter vector $\wb$---i.e., $\wb \mapsto F(\{j\in\SET{p};\ \wb_j\neq 0\})$---a structured 
sparsity-inducing norm can be built by considering its convex envelope (tightest convex lower bound) on the unit $\ell_\infty$-norm ball.
By selecting the appropriate set-function $F$, similar structures to those described above can be obtained.
This idea can be further extended to symmetric, submodular set-functions of the level sets of $\wb$, that is, $\wb \mapsto \max_{\nu\in\Real} F(\{j\in\SET{p};\ \wb_j\geq \nu\})$, thus 
leading to different types of structures~\citep{Bach2010b}, allowing to shape the level sets of $\vw$ rather than its support. This approach can also be generalized to any set-function and other priors on the the non-zero variables than the $\ell_\infty$-norm~\citep{submodlp}.

\paragraph{Non-convex approaches.}
 
We mainly focus in this review on convex penalties but in fact many non-convex approaches have been proposed as well.
In the same spirit as the norm~(\ref{eq:omegajacob}), \citet{Huang2011} considered the penalty
$$
\psi(\wb) \defin \min_{ \Hcal \subseteq \Gcal } \sum_{ g \in \Hcal }\omega_g,\ \text{such that}\ \{j\in\SET{p};\ \wb_j\neq 0\} \subseteq \bigcup_{g\in\Hcal}\!g,
$$
where $\Gcal$ is a given set of groups, and $\{\omega_g\}_{g\in\Gcal}$ is a set of positive weights which defines a \emph{coding length}. In other words,
the penalty $\psi$ measures from an information-theoretic viewpoint, ``how much it costs'' to represent $\wb$. 
Finally, in the context of compressed sensing, the work of \citet{Baraniuk2008}
also focuses on union-closed families of supports, although without
information-theoretic considerations. All of these non-convex approaches can in
fact also be relaxed to convex optimization problems~\citep{submodlp}.

\paragraph{Other forms of sparsity.}
We end this review by discussing sparse regularization
functions encoding other types of structures than the structured sparsity
penalties we have presented.
We start with the total-variation penalty
originally introduced in the image
processing community~\citep{rudin}, which encourages piecewise constant
signals. It can be found in the statistics literature under the name of
``fused lasso'' \citep{tibshirani2005sparsity}. For one-dimensional
signals, it can be seen as the~$\ell_1$-norm of finite differences for a
vector~$\vw$ in~$\R^p$:~$\Omega_{\text{TV-1D}}(\vw)\defin\sum_{i=1}^{p-1}
|\vw_{i+1}-\vw_i|$. Extensions have been proposed for multi-dimensional
signals and for recovering piecewise constant functions on
graphs~\citep{Kim2009a}.

We remark that we have presented group-sparsity penalties in Section \ref{subsec:grouplasso}, where the goal was to select a few groups of variables. A different approach called ``exclusive Lasso'' consists instead of selecting a few variables inside each group, with some applications in multitask learning~\citep{Zhou2010}.

Finally, we would like to mention a few works on automatic feature grouping~\citep{Bondell2008,Shen2010,Zhong2011}, which could be used when no a-priori group structure~$\Gcal$ is available. These penalties are typically made of pairwise terms between all variables, and encourage some coefficients
to be similar, thereby forming ``groups''.

\subsection{Convex Optimization with Proximal Methods}
\label{sec:prox}
In this section, we briefly review \emph{proximal methods} which are convex optimization methods particularly suited to the norms we have defined. They essentially allow to solve the problem regularized with a new norm at low implementation and computational costs. For a more complete presentation of optimization techniques adapted to sparsity-inducing norms, see~\citet{Bach2011}.

Proximal methods constitute a class of first-order techniques typically designed to solve problem~(\ref{eq:formulation})~\citep{Nesterov2007,Beck2009,Combettes2010}. They take advantage of the structure of~(\ref{eq:formulation}) as the sum of two convex terms. For simplicity, we will present here the proximal method known as \emph{forward-backward splitting} which assumes that at least one of these two terms,  is smooth.
Thus, we will typically assume that the loss function $\ell$ is convex differentiable, with Lipschitz-continuous gradients (such as the logistic or square loss), while $\Omega$ will only be assumed convex.

Proximal methods have become increasingly popular over the past few years, 
both in the signal processing~\citep[e.g.,][and numerous references therein]{Becker2009, Wright2009,
Combettes2010} and in the machine learning communities~\citep[e.g.,][and
references therein]{Jenatton2010b,Chen2011,Bach2011}.
In a broad sense, these methods can be described as providing a natural extension of gradient-based techniques
when the objective function to minimize has a non-smooth part.
Proximal methods are iterative procedures. 
Their basic principle is to linearize, at each iteration, the function $f$ around the current estimate
$\hat{\wb}$, and to update this estimate as the (unique, by strong convexity) solution of the so-called \textit{proximal problem}. Under the assumption that $f$ is a smooth function, it takes the form:
\begin{equation}\label{intro:eq:proxupdate}
\min_{\wb \in \Real^p}\bigg[ f(\hat{\wb}) + (\wb - \hat{\wb})^\top \nabla\! f(\hat{\wb}) + \lambda \Omega(\wb) + \frac{L}{2}\|\wb - \hat{\wb}\|_2^2 \bigg].
\end{equation}
The role of the added quadratic term is to keep the update in a neighborhood of~$\hat{\wb}$ where $f$ stays close to its current linear approximation;
 $L\!>\!0$ is a parameter which is an upper bound on the Lipschitz constant of $\nabla f$.

Provided that we can solve efficiently the proximal problem (\ref{intro:eq:proxupdate}),
this first iterative scheme constitutes a simple way of solving problem~(\ref{eq:formulation}).
It appears under various names in the literature:
proximal-gradient techniques~\citep{Nesterov2007},
forward-backward splitting methods~\citep{Combettes2010}, and
iterative shrinkage-thresholding algorithm~\citep{Beck2009}.
Furthermore, it is possible to guarantee convergence rates for the function values \citep{Nesterov2007,Beck2009},
and after $k$ iterations, the precision be shown to be of order $O(1/k)$, which should contrasted with rates for the subgradient case, that are rather $O(1/\sqrt{k})$.

This first iterative scheme can actually be extended to ``accelerated'' versions~\citep{Nesterov2007,Beck2009}. 
In that case, the update is not taken to be exactly the result from~(\ref{intro:eq:proxupdate}); instead, it is obtained as the solution of the proximal problem applied to
a well-chosen linear combination of the previous estimates.
In that case, the function values converge to the optimum with a rate of $O(1/k^2)$, where $k$ is the iteration number.
From~\citet{Nesterov2004}, we know that this rate is optimal within the class of first-order techniques; in other words, accelerated proximal-gradient methods 
can be as fast as without a non-smooth component.

We have so far given an overview of proximal methods, without specifying how we precisely handle its core part, 
namely the computation of the proximal problem, as defined in~(\ref{intro:eq:proxupdate}).

\paragraph{Proximal Problem.}
We first rewrite problem~(\ref{intro:eq:proxupdate}) as
$$
\min_{\wb \in \Real^p}~ \frac{1}{2}\Big\|\wb - \big(\hat{\wb} - \frac{1}{L}\nabla\! f(\hat{\wb}) \big) \Big\|_2^2 + \frac{\lambda}{L} \Omega(\wb).
$$
Under this form, we can readily observe that when $\lambda=0$, the solution of the proximal problem is identical to
the standard gradient update rule.
The problem above can be more generally viewed as an instance of the \textit{proximal operator}~\citep{Moreau1962} associated with $\lambda\Omega$: 
\begin{equation*}\label{intro:eq:proxoperator}
\text{Prox}_{\lambda \Omega}: \ub\in\Real^p \mapsto \argmin_{\vb \in \Real^p} \frac{1}{2} \|{\ub-\vb}\|_2^2 + \lambda  \Omega(\vb).
\end{equation*}

 For many choices of regularizers $\Omega$, the proximal problem has a closed-form solution, which makes proximal methods particularly efficient.
It turns out that for the norms defined in this paper, we can compute in a large number of cases the proximal operator exactly and efficiently \citep[see][]{Bach2011}. 
If $\Omega$ is chosen to be the $\ell_1$-norm,  the proximal operator is simply the soft-thresholding operator applied elementwise~\citep{donoho}. More formally, we have for all~$j$ in~$\SET{p}$, $\text{Prox}_{\lambda\|.\|_1}[\ub]_j = \sign(\ub_j) \max(|\ub_j|-\lambda,0)$. For the group Lasso penalty of Eq.~(\ref{intro:eq:omega}) with~$q=2$, the proximal operator is a group-thresholding operator and can be also computed in closed form: $\text{Prox}_{\lambda\Omega}[\ub]_g = (\ub_g/\|\ub_g\|_2)\max(\|\ub_g\|_2-\lambda,0)$ for all~$g$ in~$\Gcal$.
For norms with hierarchical groups of variables 
(in the sense defined in Section~\ref{sec:inter-closed}),
the computation of the proximal operator can be obtained by a composition of group-thresholding operators in a time linear in the number~$p$ of variables~\citep{Jenatton2010b}. In other settings, e.g., general overlapping groups of $\ell_\infty$-norms,
the exact proximal operator implies a more expensive polynomial dependency on $p$ using network-flow techniques~\citep{Mairal2011}, but approximate computation is possible without harming the convergence speed~\citep{Schmidt2011a}.
Most of these norms and the associated proximal problems are implemented in the open-source software SPAMS\footnote{\url{http://www.di.ens.fr/willow/SPAMS/}}.

In summary, with proximal methods, generalizing algorithms from the $\ell_1$-norm to a structured norm requires only to be able to compute the corresponding proximal operator, which can be done efficiently in many cases.

\subsection{Theoretical Analysis}
\label{sec:theory}
Sparse methods are traditionally analyzed according to three different criteria; it is often assumed that the data were generated by a sparse loading vector $\vw^\ast$. Denoting $\hat{\vw}$ a solution of the $M$-estimation problem in Eq.~(\ref{eq:formulation}), traditional statistical consistency results aim at showing that $\| \vw^\ast - \hat{\vw}\|$ is small for a certain norm $\|\cdot\|$; model consistency considers the estimation of the support of $\vw^\ast$ as a criterion, while, prediction efficiency only cares about the prediction of the model, i.e., with the square loss, the quantity $\| \Xb \vw^\ast - \Xb \hat{\vw}\|_2^2$ has to be as small as possible.

A striking consequence of assuming that $\wb^\ast$ has many zero components is that for the three criteria, consistency is achievable even when $p$ is much larger than~$n$~\citep{Zhao2006,Wainwright2009,Bickel2009,Zhang2009}. 

However, to relax the often unrealistic assumption that the data are generated by a sparse loading vector, and also because a good predictor, especially in the high-dimensional setting, can possibly be much sparser than any potential true model generating the data, prediction efficiency is often formulated under the form of \emph{oracle inequalities},
where the performance of the estimator is upper bounded by the performance of any function in a fixed complexity class, reflecting approximation error, plus a complexity term characterizing the class and reflecting the hardness of estimation in that class. We refer the reader to \citet{Geer2010L1} for a review and references on oracle results for the Lasso and the group Lasso.

It should be noted that model selection consistency and prediction efficiency are obtained in quite different regimes of regularization, so that it is not possible to obtain both types of consistency with the same Lasso estimator \citep{Shalev2010Trading}.
For prediction consistency, the regularization parameter is easily chosen by cross-validation on the prediction error.
For model selection consistency, the regularization coefficient should typically be much larger than for prediction consistency; but rather than trying to select an optimal regularization parameter in that case, it is more natural to
consider the collection of models obtained along the regularization path and to apply usual model selection methods
to choose the best model in the collection. One method that works reasonably well in practice, sometimes called ``OLS hybrid'' for the least squares loss \citep{Efron2004}, consists in refitting the different models without regularization and to choose the model with the best fit by cross-validation.

In structured sparse situations, such high-dimensional phenomena can also be characterized. Essentially, if one can make the assumption that $\vw^\ast$ is compatible with the additional prior knowledge on the sparsity pattern encoded in the norm~$\Omega$, then, some of the assumptions required for consistency can sometimes be relaxed \citep[see][]{Huang2010,Jenatton2009,Huang2011,Bach2010a}, and faster rates can sometimes be obtained \citep{Huang2010,Huang2011,Obozinski2011Support,Negahban2011Simultaneous,hkl,Percival2011}. 
However, one major difficulty that arises is that some of the conditions for recovery or to obtain fast rates of convergence depend on an intricate interaction between the sparsity pattern, the design matrix and the noise covariance, which leads in each case to sufficient conditions that are typically not directly comparable between different structured or unstructured cases~\citep{Jenatton2009}. Moreover, even if the sufficient conditions are satisfied simultaneously for the norms to be compared, sharper bounds on rates and sample complexities would still often be needed to characterize more accurately the improvement resulting from having a stronger structural a priori.

\section{Sparse principal component analysis and dictionary learning}
\label{sec:unsup}
Unsupervised learning aims at extracting latent representations of the data that are useful for analysis, visualization, denoising
or to extract relevant information to solve subsequently a supervised learning problem.
Sparsity or structured sparsity are essential to specify, on the representations, constraints that improve their identifiability and interpretability.

\subsection{Analysis and Synthesis Views of PCA}
Depending on how the latent representation is extracted or constructed from the data, it is useful to distinguish two points of view. 
This is illustrated well in the case of PCA.

In the \emph{analysis} view, PCA aims at finding \emph{sequentially} a set of directions in space
that explain the largest fraction of the variance of the data. This can be formulated as an iterative procedure
in which a one-dimensional projection of the data with maximal variance is found first, then the data are projected on the orthogonal subspace 
(corresponding to a \emph{deflation} of the covariance matrix),
and the process is iterated.
In the \emph{synthesis} view, PCA aims at finding a set of vectors, or \emph{dictionary elements} (in a terminology closer to signal processing)
such that all observed signals admit a linear decomposition on that set with low
reconstruction error. 
In the case of PCA, these two formulations lead to the same solution (an eigenvalue problem). However, in extensions of PCA, in which either the dictionary elements or
the decompositions of signals are constrained to be sparse or structured, they lead to different algorithms with different solutions.

The \emph{analysis} interpretation leads to sequential formulations \citep{dAspremont2008,Moghaddam2006,Jolliffe2003}
that consider components one at a time and perform a \emph{deflation}  of the
covariance matrix at each step \citep[see][]{Mackey2009}. The \emph{synthesis}
interpretation leads to non-convex global formulations \citep[see, e.g.,][]{Zou2006a,Moghaddam2006,Aharon2006,Mairal2010} which estimate
simultaneously all principal components, typically do not require the orthogonality of the components, 
and are referred to as matrix factorization problems \citep{Singh2008,Bach2008c} in
machine learning, and dictionary learning in signal processing~\citep{Olshausen1996}.

While we could also impose structured sparse priors in the analysis view, we will consider from now on the synthesis view, that we will introduce with the terminology of {dictionary learning}.

 \subsection{Dictionary Learning}
Given a matrix $\Xb \in \Real^{m\times n}$ of $n$ columns corresponding to $n$ observations in $\Real^{m}$, 
the dictionary learning problem is to find a matrix $\Db \in \Real^{m\times p}$, called \emph{dictionary}, 
such that each observation can be well approximated by a linear combination of the $p$ columns $(\db^k)_{k \in \SET{p}}$ of $\Db$ called the \emph{dictionary elements}. 
If $\Ab \in \Real^{p \times n}$ is the matrix of the linear combination coefficients or \emph{decomposition coefficients} (or \emph{codes}), with $\alphab^k$ the $k$-th column of $\Ab$ being the coefficients for the $k$-th signal $\xb^k$, the matrix product $\Db\Ab$ is called a decomposition of~$\Xb$.

Learning simultaneously the dictionary $\Db$ and the coefficients $\Ab$ corresponds to a matrix factorization problem \citep[see][and reference therein]{Witten2009}. 

As formulated by \citet{Bach2008c} or \citet{Witten2009}, it is natural, when learning a decomposition, to penalize or constrain some norms or pseudo-norms of $\Ab$ and $\Db$, 
say  $\Omega_\Ab$ and $\Omega_\Db$ respectively, to encode prior information --- typically sparsity --- about the decomposition of~$\Xb$. While in general the penalties could be defined globally on the matrices $\Ab$ and $\Db$, we assume that each column of $\Db$ and $\Ab$ is penalized separately. This can be written as

\begin{equation}\label{sspca:eq:main_eq}
      \min_{  \substack{\Ab \in \Real^{p\times n},\\ \Db \in \Real^{m\times p}}  }\ \frac{1}{2nm} \big\| \Xb \! - \! \Db \Ab  \big\|^2_\fro + \lambda \sum_{k=1}^p \Omega_\Db(\db^k),\
      \text{s.t.}\ \Omega_\Ab(\alphab^i)\, \leq \, 1,\ \forall\, i\in\SET{n},
\end{equation}
where the regularization parameter $\lambda \geq 0$ controls to which extent the dictionary is regularized.
If we assume that both regularizations $\Omega_\Ab$ and $\Omega_\Db$ are convex, problem (\ref{sspca:eq:main_eq})  
is convex with respect to~$\Ab$ for fixed $\Db$ and vice versa. It is however not \emph{jointly} convex in the pair $(\Ab,\Db)$, but alternating optimization schemes generally lead to good performance in practice.

\subsection{Imposing Sparsity}
The choice of the two norms $\Omega_\Ab$ and $\Omega_\Db$  is crucial and heavily influences the behavior of dictionary learning. Without regularization, any solution $(\Db,\Ab)$ is such that $\Db\Ab$ is the best fixed-rank approximation of~$\Xb$, and the problem can be solved exactly with a classical PCA.
When
$\Omega_\Ab$ is the $\ell_1$-norm and $\Omega_\Db$ the $\ell_2$-norm, we aim at finding a dictionary such that each signal $\vx^i$ admits a sparse decomposition on the dictionary. In this context, we are essentially looking for a basis where the data have sparse decompositions, a framework we refer to as \emph{sparse dictionary learning}. On the contrary,  when
$\Omega_\Ab$ is the $\ell_2$-norm and $\Omega_\Db$ the $\ell_1$-norm, the formulation induces sparse principal components, i.e., atoms with many zeros, a framework we refer to as sparse PCA. In \mysec{sspca} and \mysec{hdl}, we replace the $\ell_1$-norm by structured norms introduced in \mysec{structsparsity}, leading to structured versions of the above estimation problems.

\subsection{Adding Structures to Principal Components}
\label{sec:sspca}

One of PCA's main shortcomings is that, even if it finds a small number of
important factors, the factor themselves typically involve all original
variables.
In the last decade, several alternatives to PCA which find sparse and potentially interpretable factors have been proposed, 
notably non-negative matrix factorization (NMF) \citep{Lee1999} 
and sparse PCA (SPCA) \citep{Jolliffe2003,Zou2006a,Zass2007,Witten2009}.  

However, in many applications, only constraining the size of the supports of the factors
does not seem appropriate because the considered factors are not only expected
to be sparse but also to have a certain structure.
In fact, the popularity of NMF for face image analysis owes essentially to
the fact that the method happens to retrieve sets of variables that are partly
localized on the face and capture some features or parts of the face which
seem intuitively meaningful given our a priori. We might therefore gain in
the quality of the factors induced by enforcing directly this a priori in
the matrix factorization constraints. More generally, it would be desirable to
encode higher-order information about the supports that reflects the
\textit{structure} of the data. For example, in computer vision, features
associated to the pixels of an image are naturally organized on a grid and
the supports of factors explaining the variability of images could be
expected to be localized, connected or have some other regularity with
respect to that grid. Similarly, in genomics, factors explaining the gene
expression patterns observed on a microarray could be expected to involve
groups of genes corresponding to biological pathways or set of genes that
are neighbors in a protein-protein interaction network.

Based on these remarks and with the norms presented earlier, sparse PCA is readily extended to \textit{structured sparse PCA} (SSPCA),
which explains the variance of the data by factors that are not only sparse but
also respect some a priori structural constraints deemed relevant to model
the data at hand: slight variants of the regularization term
defined in \mysec{structsparsity} (with the groups defined in Figure~\ref{intro:fig:axis-aligned}) can be used successfully for $\Omega_\Db$.

\paragraph{Experiments on face recognition.}

By definition, dictionary learning belongs to unsupervised learning; in that sense, our method may appear first as a tool for exploratory data analysis,
which leads us naturally to \emph{qualitatively} analyze the results of our decompositions (e.g., by visualizing the learned dictionaries). This is obviously a difficult and subjective exercise, beyond the assessment of the consistency of the method in artificial examples where the ``true'' dictionary is known. For quantitative results, see \citet{Jenatton2010}.\footnote{A Matlab toolbox implementing our method can be downloaded from {\small \url{http://www.di.ens.fr/~jenatton/}}.}

We apply SSPCA on the cropped AR Face Database \citep{Martinez2001} that consists of 2600 face images, corresponding to 100 individuals (50 women and 50 men). For each subject, there are 14 non-occluded poses and 12 occluded ones (the occlusions are due to sunglasses and scarfs). We reduce the resolution of the images from $165\!\times\! 120$ pixels to $38 \!\times\! 27$ pixels for computational reasons.

\begin{figure}[!h]
	\vspace*{-0.2cm}
	\begin{center}
	\begin{tabular}{cc}
	\includegraphics[scale=0.52]{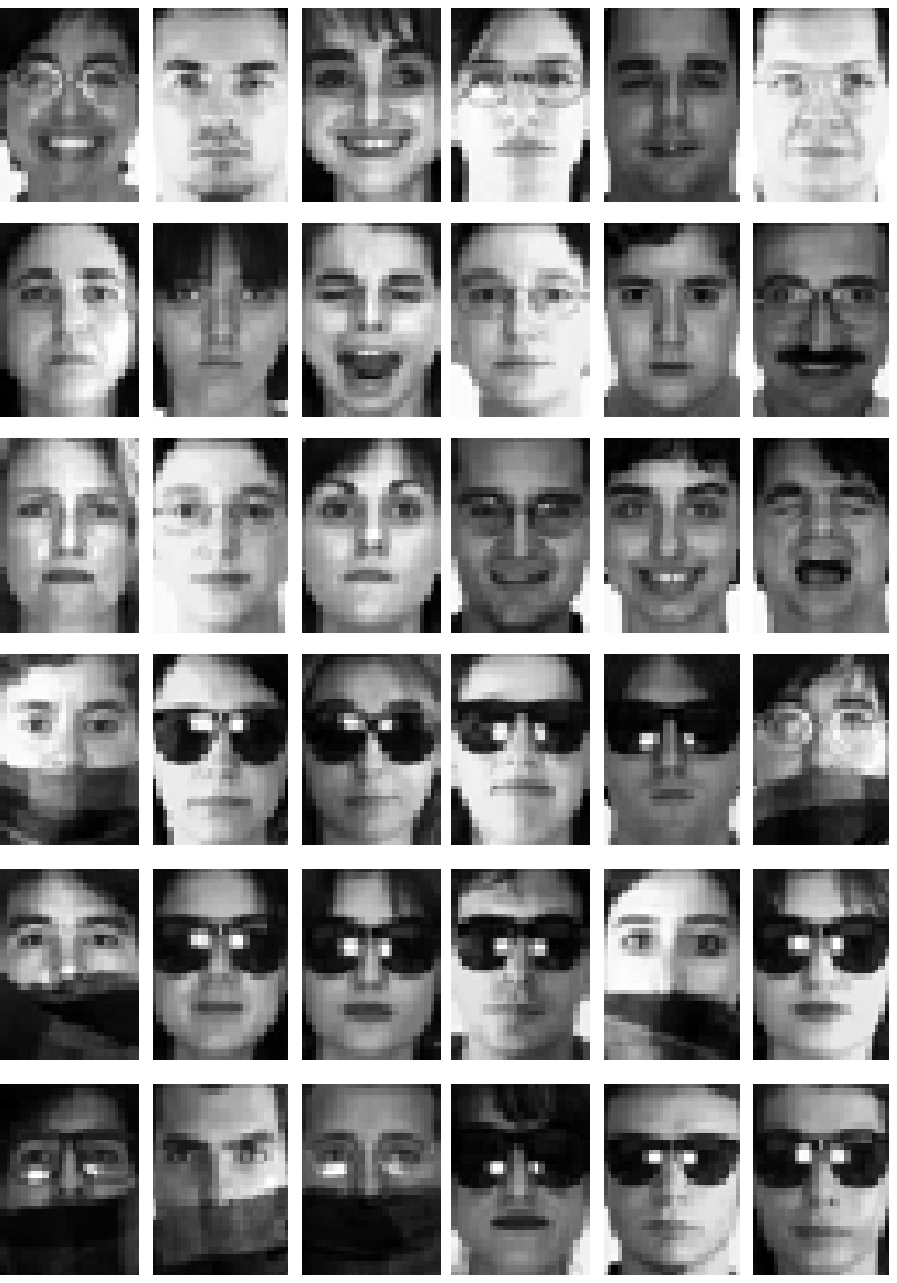}  &  \includegraphics[scale=0.52]{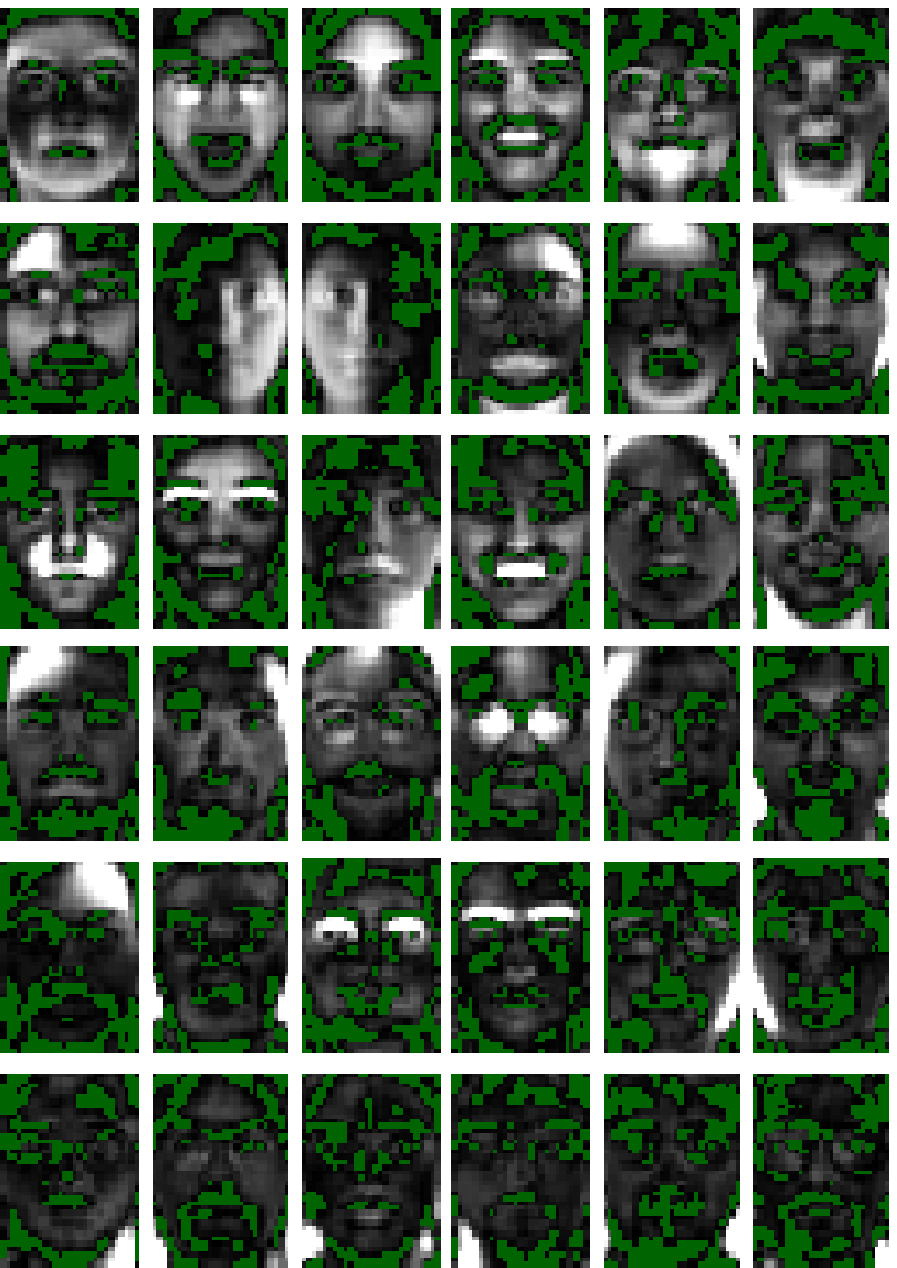} \\ 
	\includegraphics[scale=0.52]{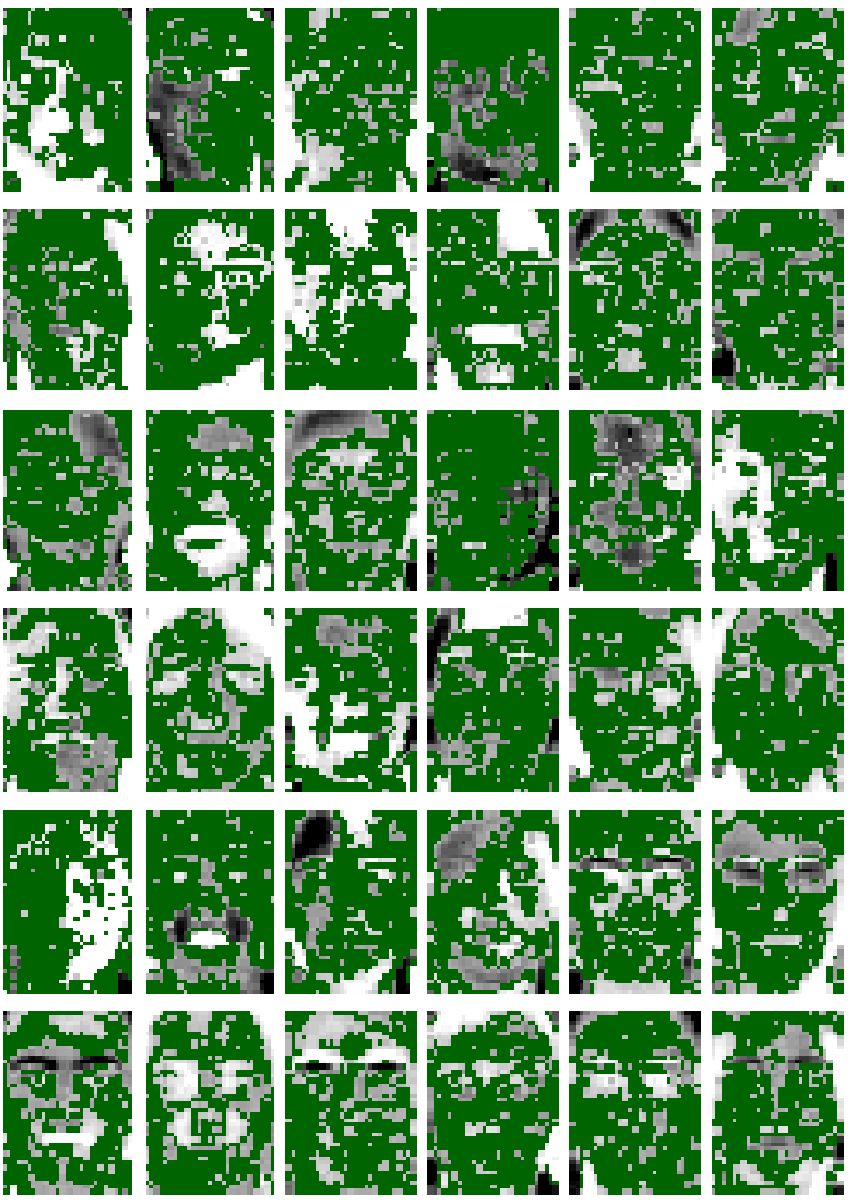}      &  \includegraphics[scale=0.52]{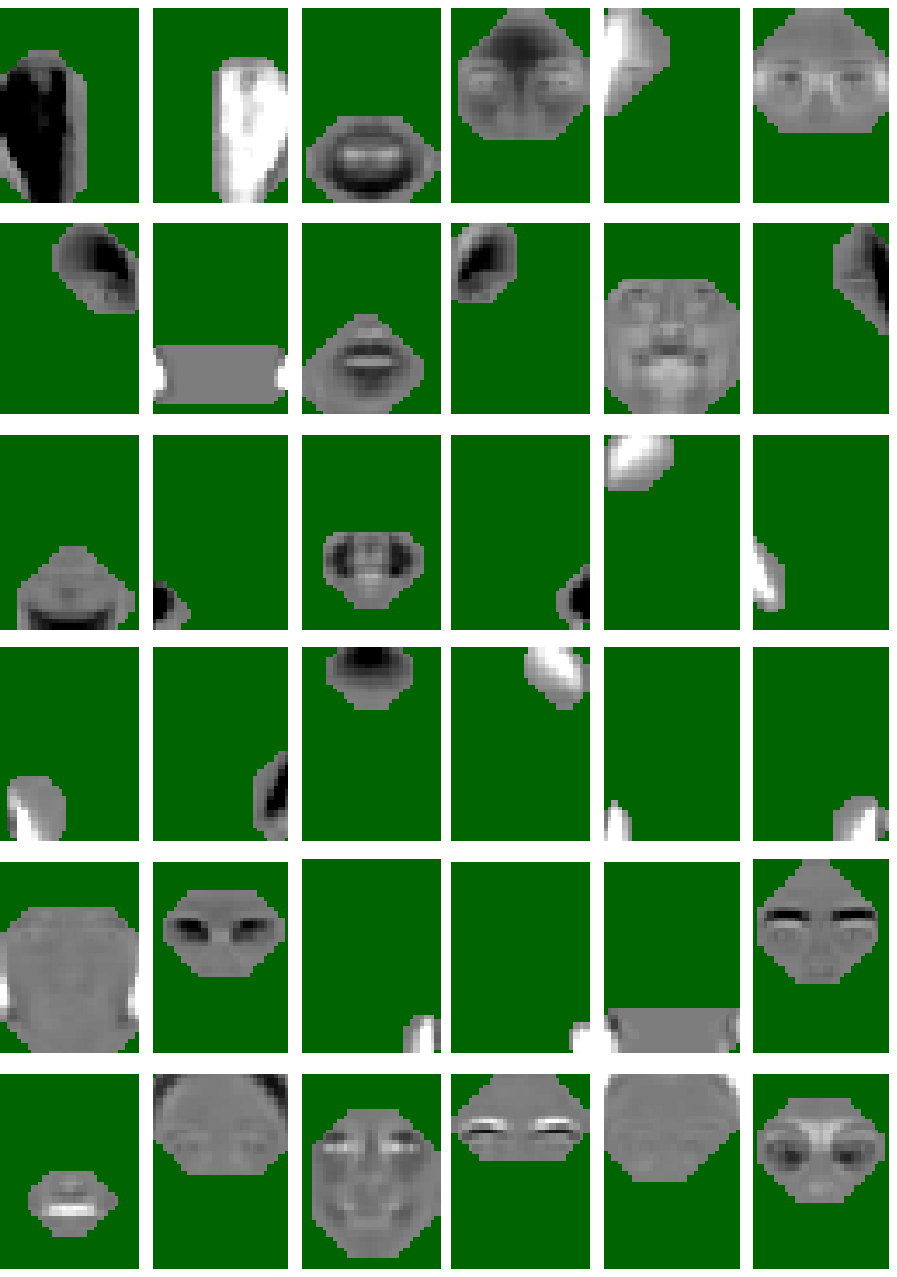}
	\end{tabular}
	\end{center}
	\vspace*{-0.2cm}
\caption{ Top left, examples of faces in the datasets. The three remaining images represent learned dictionaries of faces with $p\!=\!36$:  
NMF (top right), SPCA (bottom left) and SSPCA (bottom right). 
The dictionary elements are sorted in decreasing order of explained variance. While NMF gives sparse spatially unconstrained patterns, SSPCA finds convex areas that correspond to more natural face segments. SSPCA captures the left/right illuminations and retrieves pairs of symmetric patterns.
Some displayed patterns do not seem to be convex, e.g., nonzero patterns located at two opposite corners of the grid. 
However, a closer look at these dictionary elements shows that convex shapes are indeed selected, and that small numerical values 
(just as regularizing by $\ell_2$-norm may lead to) give the visual impression of having zeroes in convex nonzero patterns.
This also shows that if a nonconvex pattern has to be selected, 
it will be, by considering its convex hull.}
\label{sspca:fig:face_dictionary_examples}
\end{figure}

Figure~\ref{sspca:fig:face_dictionary_examples} shows examples of learned dictionaries (for $p\!=\!36$ elements), 
for NMF, unstructured sparse PCA (SPCA), and SSPCA. While NMF finds sparse but spatially unconstrained patterns, SSPCA selects sparse convex areas that correspond to a more natural segment of faces. For instance, meaningful parts such as the mouth and the eyes are recovered by the dictionary.

\subsection{Hierarchical Dictionary Learning}
\label{sec:hdl}
In this section, we consider sparse dictionary learning, where the structured sparse prior knowledge is put on the decomposition coefficients, i.e., the matrix~$\Ab$ in Eq.~(\ref{sspca:eq:main_eq}),
and present an application to text documents.

\paragraph{Text documents.}\label{prox:sec:exp_txt_documents}
The goal of probabilistic topic models is to find a low-dimen\-sio\-nal representation
of a collection of documents, where
the representation should provide a semantic description of the collection.
Approaching the problem in a parametric Bayesian framework,
latent Dirichlet allocation (LDA), \citet{Blei2003} models documents, represented as vectors of word counts,
as a mixture of a predefined number of \emph{latent topics}, defined as multinomial distributions over a fixed vocabulary. 
The number of topics is usually small compared to the size of the vocabulary (e.g., 100 against $10\, 000$),
so that the topic proportions of each document provide a compact representation of the corpus.


In fact the problem addressed by LDA is fundamentally a matrix factorization problem. For instance, \citet{Buntine2002} argued that LDA 
can be interpreted as a Dirichlet-multinomial counterpart of factor analysis.
We can actually cast the problem in the dictionary learning formulation that we presented before\footnote{Doing so we simply trade the multinomial likelihood with a least-square formulation.}.
Indeed, suppose that the signals
$ \Xb = [\xb^1,\dots,\xb^n]$ in $\Real^{m\times n}$
are each the so-called \emph{bag-of-word} representation of each of $n$ documents over a vocabulary of $m$ words, i.e., $\xb^i$ is a vector whose 
$k$-th component is the empirical frequency in document~$i$ of the $k$-th word of a fixed lexicon.
If we constrain the entries of $\Db$ and $\Ab$ to be nonnegative,
and the dictionary elements~$\db^j$ to have unit $\ell_1$-norm,
the decomposition~$(\Db,\Ab)$ can be interpreted as the parameters of a topic-mixture model.
Sparsity here ensures that a document is described by a small number of topics.

Switching to structured sparsity allows in this case to organize automatically the dictionary of topics in the process of learning it.
Assume that $\Omega_\Ab$ in Eq.~(\ref{sspca:eq:main_eq}) is a tree-structured regularization, such as illustrated on Figure~\ref{intro:fig:treegroups}; in this case, in the light of \mysec{inter-closed}, if the decomposition of a document involves a certain topic, then all ancestral topics in the tree are also present in
the topic decomposition.
Since the hierarchy is shared by all documents, the topics close to the root
participate in every decomposition, and given that the dictionary is learned, this mechanism forces those topics 
to be quite generic---essentially gathering the lexicon which is common to all documents.
Conversely, the deeper the topics in the tree, the more specific they should be.
It should be noted that such hierarchical dictionaries can also be obtained with generative probabilistic models, typically based on non-parametric Bayesian prior over trees or paths in trees, and which extend the LDA model to topic hierarchies \citep{Blei2010,Adams2010Tree}.
\begin{figure}[h!]
\centering
\includegraphics[width=0.65\linewidth]{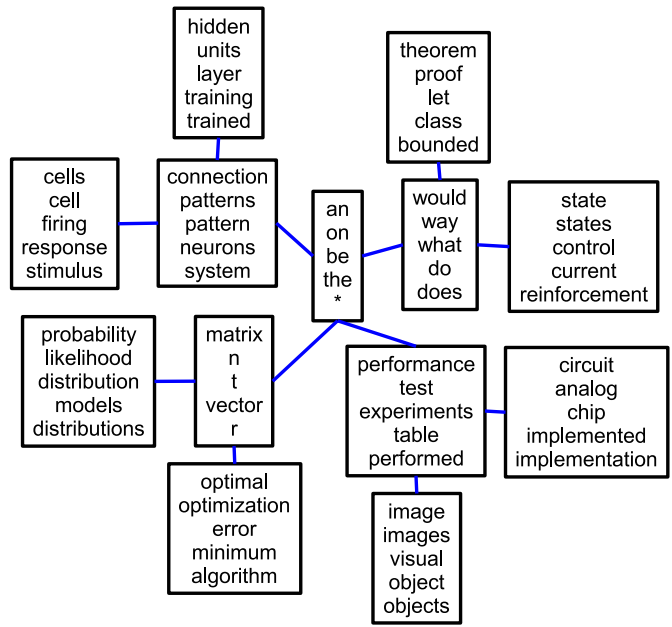}
\caption{Example of a topic hierarchy estimated from 1714 NIPS proceedings papers (from 1988 through 1999).
Each node corresponds to a topic whose 5 most important words are displayed.
Single characters such as $n, t, r$ are part of the vocabulary and often appear in NIPS papers, and their place in the hierarchy is semantically relevant to children topics.}
\label{prox:fig:topics_nips}
\end{figure}

\paragraph{Visualization of NIPS proceedings.}
We qualitatively illustrate our approach on the NIPS proceedings
from 1988 through 1999 \citep{Griffiths2004}.
After removing words appearing fewer than 10 times, the dataset is composed of 1714 articles, with
a vocabulary of 8274 words.
As explained above, we enforce both the dictionary and the sparse coefficients to be non-negative,
and constrain the dictionary elements to have a unit~$\ell_1$-norm.
Figure~\ref{prox:fig:topics_nips} displays an example of a learned
dictionary with 13 topics,
obtained by using a tree-structured penalty (see Section~\ref{sec:inter-closed}) on the coefficients $\Ab$ and by selecting manually\footnote{The regularization parameter striking a good compromise between sparsity and reconstruction of the data is chosen here by hand because (a) cross-validation would yield a significantly less sparse dictionary and (b) model selection criteria would not apply without serious caveats here since the dictionary is learned at the same time.} $\lambda\!=\!2^{-15}$.
As expected and similarly to~\citet{Blei2010}, we capture the stopwords at the root of the tree, and topics reflecting the different subdomains of the conference
such as neurosciences, optimization or learning theory.
 
\section{High-dimensional non-linear variable selection}
\label{sec:hkl}

In this section, we show how structured sparsity-inducing norms may be used to provide an efficient solution to the problem of high-dimensional \emph{non-linear} variable selection. Namely, given $p$ variables $\vx=(\vx_1, \dots,\vx_p)$, our aim is to find a non-linear function $\fb(\vx_1,\dots,\vx_p)$ which depends only on a few variables. First approaches to the problem have considered restricted functional forms such as $\fb(\vx_1,\dots,\vx_p) = \fb_1(\vx_1) + \cdots + \fb_p(\vx_p)$, where each $\fb_1,\dots,\fb_p$ are univariate non-linear functions~\citep{ravikumar2009sparse,Bach2008a}. However, many non-linear functions cannot be expressed as sums of functions of these forms. Additional interactions have been added leading to functions of the form $  \fb(\vx_1,\dots,\vx_p) = \sum_{J \subset \{1,\dots, p\}, \ |J| \leqslant 2} \fb_J(\vx_J)$~\citep{cosso}. While second-order interactions make the class of functions larger, our aim in this section is to consider functions which can be expressed as a sparse linear combination of the form
$  \fb(\vx_1,\dots,\vx_p) = \sum_{J \subset \{1,\dots, p\} } \fb_J(\vx_J)$, i.e., a combination of functions defined on potentially larger subsets of variables.

The main difficulties associated with this problem are that (1) each function~$\fb_J$ has to be estimated, leading to a non-parametric problem, and (2) there are exponentially many such functions. We propose however an approach that overcomes both difficulties in the next section, based on the ideas that estimating functions rather than vectors can be tackled with estimators in reproducing kernel Hilbert spaces (see \mysec{mkl}), and that the complexity issues can be addressed by imposing some structure among all the subsets $J \subset \{1,\dots,p\}$ (see \mysec{hkl}).

\subsection{Multiple Kernel Learning: From Linear to Non-Linear Predictions}
\label{sec:mkl}

Reproducing kernel Hilbert spaces are arguably the simplest spaces for the non-parametric estimation of non-linear functions since most learning algorithms for linear models are directly ported to any RKHS via simple kernelization. 
We therefore start by reviewing learning from a single and later multiple reproducing kernels, since our approach will be based on combining functions from multiple (actually a hierarchy) of RKHSes. For more details, see~\citet{Bach2008a}.

\paragraph{Single kernel learning.}
Let us assume that the $n$ input  data-points $\vx^{(1)},\dots,\vx^{(n)}$ belong to a set $\Xc$ (not necessarily $\Real^p$), and consider predictors of the form $\langle f, \Phi(\vx) \rangle$ where $\Phi: \Xc \to \mathcal{F}$ is a map from the input space to a reproducing kernel Hilbert space $\mathcal{F}$ (associated to the kernel function $k$), which we refer to as the feature space. These predictors are linearly parameterized, but may depend non-linearly on $\vx$. We consider the following estimation problem:
$$
\min_{ \fb \in \mathcal{F} } \frac{1}{n}\sum_{i=1}^n\ell(y^{(i)}, \langle \fb, \Phi(\vx^{(i)}) \rangle ) + \frac{\lambda}{2} \| \fb \|_\mathcal{F}^2,
$$
where~$\|.\|_\mathcal{F}$ is the Hilbertian norm associated to~$\mathcal F$.
The representer theorem~\citep{representer} states that, for all loss functions (potentially nonconvex), the solution $\fb$ admits the expansion $\fb = \sum_{i=1}^n \alphab_i \Phi(\vx^{(i)})$, so that, replacing $\fb$ by its new expression, we can now minimize
$$
\min_{ \alphab \in \Real^n } \frac{1}{n}\sum_{i=1}^n\ell(y^{(i)},  ( \Kb \alphab)_i ) + \frac{\lambda}{2} \alphab^\top \Kb \alphab,
$$
where $\Kb$ is the \emph{kernel matrix}, an $n \times n$ matrix whose element $(i,j)$ is equal to $\langle \Phi(\vx^{(i)}), \Phi(\vx^{(j)}) \rangle = k(\vx^{(i)},\vx^{(j)})$.
This optimization problem involves the observations $\vx^{(1)},\dots,\vx^{(n)}$ only through the kernel matrix~$\Kb$, and can thus be solved, as long as $\Kb$ can be evaluated efficiently. See~\cite{Shawe-Taylor2004} for more details.

\paragraph{Multiple kernel learning (MKL).}
We can now assume that we are given $m$ Hilbert spaces~$\mathcal{F}_j$, $j=1,\dots, m$, and look for predictors of the form
$\fb(\vx) = \gb_1(\vx) + \cdots + \gb_m(\vx)$, where\footnote{Notice that the function $\gb_j$ is not restricted to depend only on a subpart of $\xb$ as before.} each $\gb_j \in \mathcal{F}_j$. In order to have many $\gb_j$ equal to zero, we can penalize $\fb$ using a sum of norms similar to the group Lasso penalties introduced earlier, namely
$\sum_{j=1}^m \|\gb_j \|_{\mathcal{F}_j}$. This leads to selection of functions.
Moreover, it turns out that the optimization problems may be expressed also in terms of the $m$ kernel matrices, and it is equivalent to learn a sparse linear combination $\hat{\Kb}=\sum_{j=1}^m \eta_j \Kb_j$ (with many $\eta_j$'s equal to zero) of kernel matrices with then $\alphab$ solution of the single kernel learning problem for  $\hat{\Kb}$.
For more details, see~\citet{Bach2008a}.

\paragraph{From MKL to sparse generalized additive models.}
As shown above, the MKL framework is defined with any set of $m$ RKHSes defined on the same base set~$\mathcal{X}$. When the base set is itself defined as a cartesian product of $p$ base sets, i.e., $\mathcal{X} = \mathcal{X}_1 \times 
\cdots \times \mathcal{X}_p$, then it is common to consider $m=p$ RKHSes which are each of them defined on a single $\mathcal{X}_i$, leading to the desired functional form $\fb_1(\vx_1) + \cdots + \fb_p(\vx_p)$. To overcome the limitation of this functional form we need to consider a more complex expansion.

\subsection{Hierarchical Kernel Learning}
\label{sec:hkll}


In this section, we consider functional expansions with up to $m=2^p$ terms corresponding to different RKHSes, each defined on a cartesian product of a subset of the $p$ separate input spaces.
Specifically, we consider functions of the form $\fb(\vx_1,\dots,\vx_p) = \sum_{J \subset \{1,\dots, p\} } \fb_J(\vx_J)$ with $\fb_J$ chosen to live in a RKHS $\mathcal{F}_J$ defined on variables $\vx_J$. Penalizing by the norm $\sum_{J \subset \{1,\dots,p\}} \|\fb_J \|_{\mathcal{F}_J}$ would in theory lead to an appropriate selection of functions from the various RKHSes (and to learning a sparse linear combination of the corresponding kernel matrices). However, in practice, there are $2^p$ such predictors, which is not algorithmically feasible. 

\begin{figure}
\begin{center}
\hspace*{-1.75cm}
\includegraphics[scale=.43]{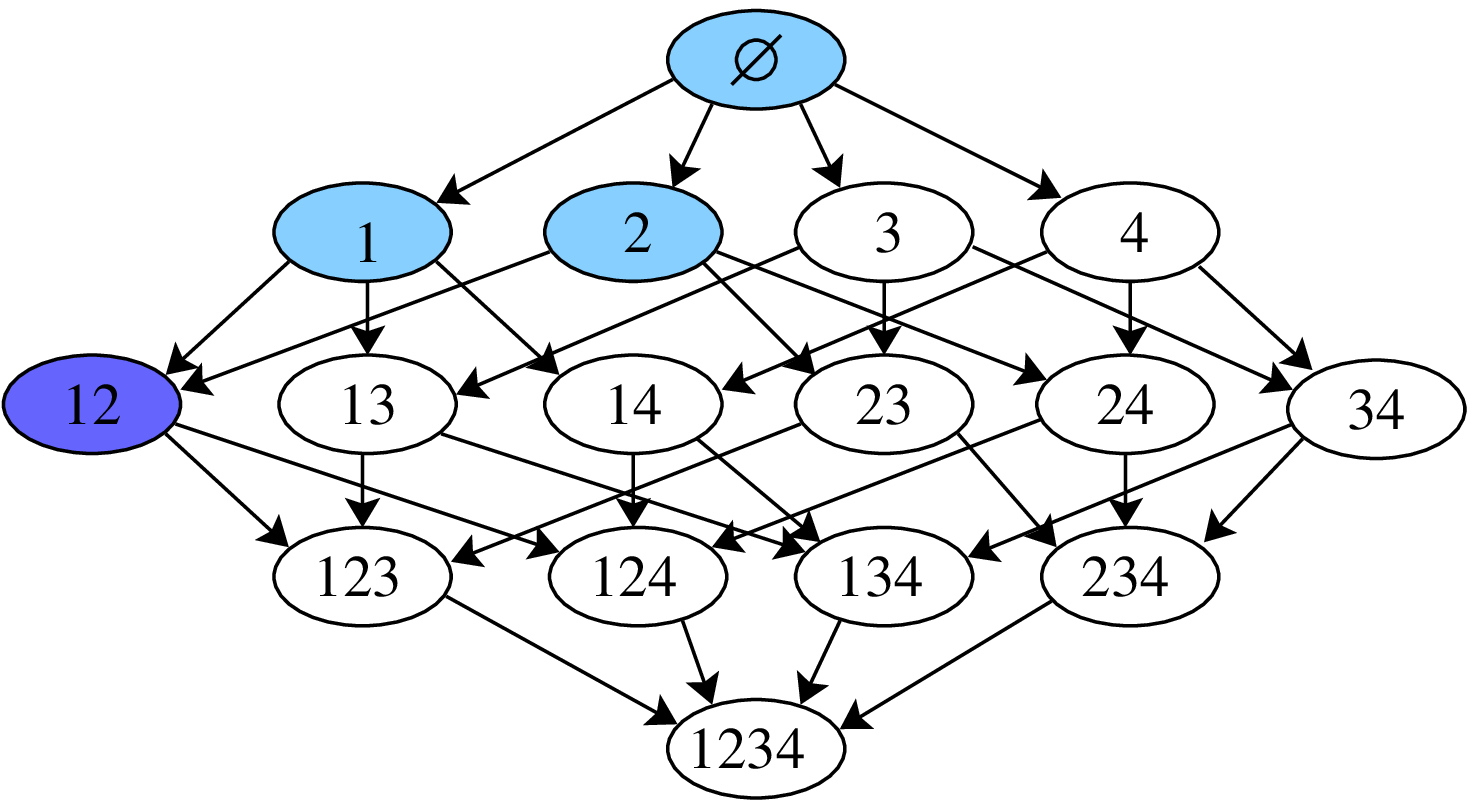} \hspace*{.1cm}
\includegraphics[scale=.43]{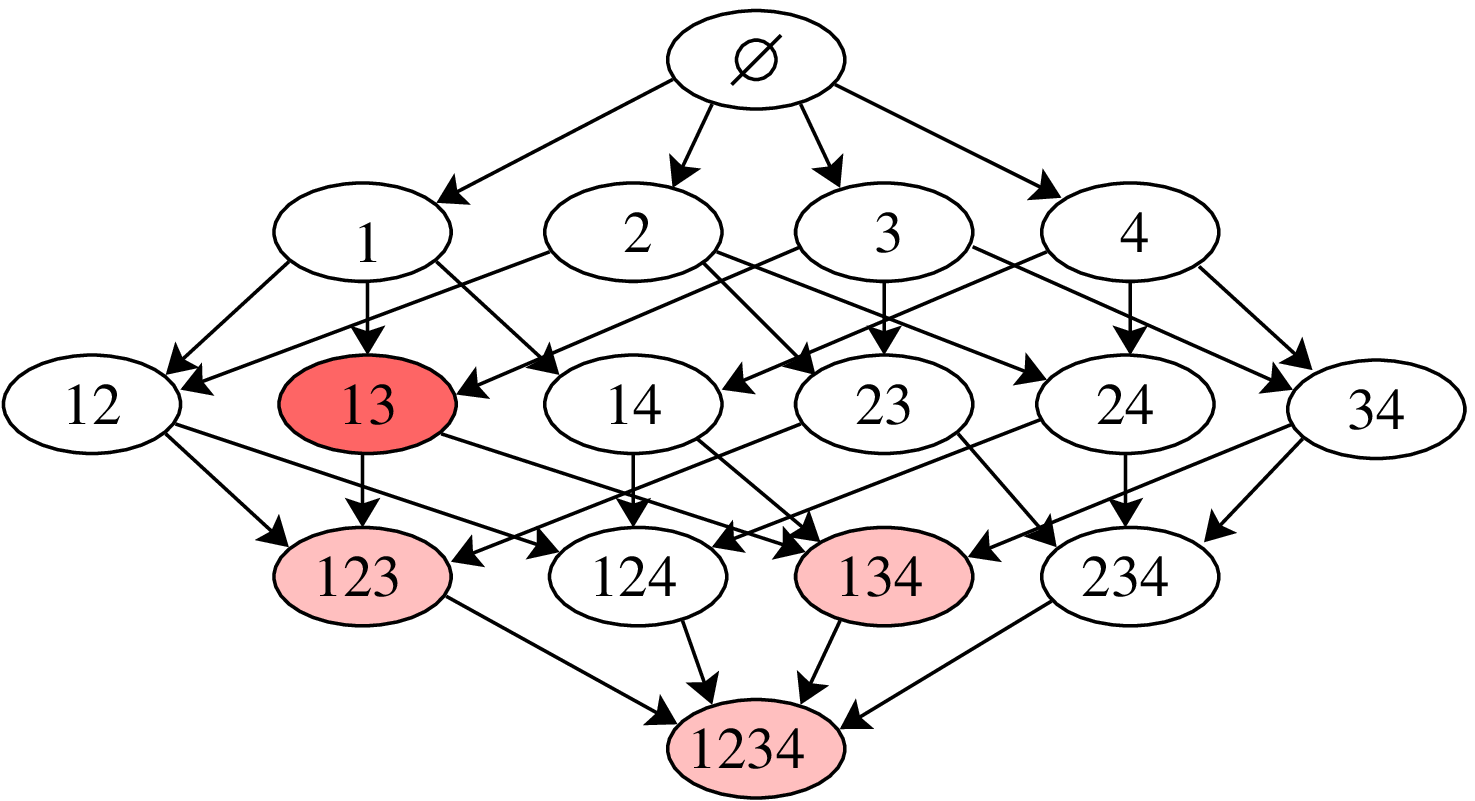}
\hspace*{-2cm}
\caption{Power set of the set $\{1,\dots,4\}$: in blue, an authorized set of selected subsets. In red, an example of a group used within the norm (a subset and all of its descendants in the DAG).
\label{fig:hkl}}
\end{center}

 \end{figure}

This is where structured sparsity comes into play. In order to obtain polynomial-time algorithms and theoretically controlled predictive performance, we may add an extra constraint to the problem.
Namely, we endow the power set of $\{1,\dots,p\}$ with the partial order of the inclusion of sets, and in this directed acyclic graph (DAG), we require that predictors $\fb$ select a subset only after all of its ancestors have been selected.
This can be achieved in a convex formulation using a structured-sparsity inducing norm of the type presented in Section~\ref{sec:inter-closed}, but defined by a hierarchy of groups as follows
$$ \Omega \big [ (\fb_H\big)_{H \subset \{1, \ldots, p\}} ] = \sum_{J \subset \{1,\dots,p\} }  \bigg( \sum_{H \supset J } \| \fb_H \|_2^2  \bigg)^{1/2}.$$
As illustrated in Figure~\ref{fig:hkl}, this norm corresponds to overlapping groups of variables defined on the directed acyclic graphs of all subsets of $\{1,\dots,p\}$.
We will explain briefly how introducing this norm may lead to polynomial time algorithms and what theoretical guarantees are associated with it.
Illustrations of the application of hierarchical kernel learning to real data can be found in~\citet{hkl}.

\paragraph{Polynomial-time estimation algorithm.}
While we are, a priori, still facing an estimation problem with $2^p$ functions, it can be solved using an active set method, which considers adding a component $\fb_J \in \mathcal{F}_J$ (resp. $\Kb_J$) to the active set of predictors (resp. kernels). The two crucial aspects are (1) to add the right kernel, i.e., choose among the $2^p$ which one to add, and (2) when to stop. As shown in \cite{hkl}, these steps may be carried out efficiently for certain collections of RKHSes $\mathcal{F}_J$, in particular those for which we are able to compute efficiently (i.e., in polynomial time in $p$) the sum $\sum_{J \subset \{1,\dots,p\}} \Kb_J$. This is the case, for example, for Gaussian kernels $k_J(\vx_J,\vx_J') = 
\exp( - \gamma \| \vx_J - \vx_J' \|_2^2` )$.

\paragraph{Theoretical analysis.}

\hspace{1mm}\cite{hkl} showed that under appropriate assumptions, estimation under high-dimensional scaling, i.e., for $p \gg n$ but $ \log p = O(n)$, is possible in this situation, in spite of the fact that the number of terms in the expansion is now potentially doubly exponential in $n$. 
 
\section{Conclusion}
In this paper, we reviewed several approaches for structured sparsity, based on convex optimization and the design of appropriate sparsity-inducing norms. Analyses and algorithms for the traditional $\ell_1$-norm can readily be extended to these new norms, making them an efficient and flexible tools for introducing prior knowledge in high-dimensional statistical problems.
We also presented several applications to supervised and unsupervised learning problems, where the proper use of additional knowledge leads to improved interpretability of the sparse estimates and/or increased predictive performance.

\label{sec:conclusion}

\section*{Acknowledgements}
Francis Bach, Rodolphe Jenatton and Guillaume Obozinski are supported in
part by ANR under grant MGA ANR-07-BLAN-0311 and the European Research
Council (SIERRA Project).
Julien Mairal is supported by the NSF grant SES-0835531 and NSF award
CCF-0939370.
The authors would like to thank the anonymous reviewers, whose comments have greatly contributed to improve the quality of this paper.

\bibliographystyle{imsart-nameyear}
\bibliography{bach_statscience}

\end{document}